Master Thesis

IU University of Applied Sciences
Study program: M.S. Artificial Intelligence

# Explicit User Manipulation in Reinforcement Learning Based Recommender Systems

Matthew Sparr

January 17, 2022

# Acknowledgments

I would like to thank my professor and supervisor, Dr. Max Pumperla for his advice and guidance on this thesis project. Several brainstorming sessions lead me to a topic in which I became engrossed. I would also like to thank those in my professional and personal life for their continued support throughout the process.

# Abstract

Recommender systems are highly prevalent in the modern world due to their value to both users and platforms and services that employ them. Generally, they can improve the user experience and help to increase satisfaction, but they do not come without risks. One such risk is that of their effect on users and their ability to play an active role in shaping user preferences. This risk is more significant for reinforcement learning based recommender systems. These are capable of learning for instance, how recommended content shown to a user today may tamper that user's preference for other content recommended in the future. Reinforcement learning based recommendation systems can thus implicitly learn to influence users if that means maximizing clicks, engagement, or consumption.

On social news and media platforms, in particular, this type of behavior is cause for alarm. Social media undoubtedly plays a role in public opinion and has been shown to be a contributing factor to increased political polarization. Recommender systems on such platforms, therefore, have great potential to influence users in undesirable ways. However, it may also be possible for this form of manipulation to be used intentionally. With advancements in political opinion dynamics modeling and larger collections of user data, explicit user manipulation in which the beliefs and opinions of users are tailored towards a certain end emerges as a significant concern in reinforcement learning based recommender systems.

# Table of Contents

# List of Figures

# 1| Introduction

## 1.1 Motivation

Found across nearly all realms of the internet, recommender systems are nowadays ubiquitous. From movies and music to shopping and news, these algorithms seek to personalize the user experience by recommending content the user is most likely to consume, engage with, or simply enjoy [42]. Sometimes this is obvious to the user – for example, a "recommended for you" queue on a streaming service – but sometimes it is not obvious and is working more behind-the-scenes.

Reinforcement learning based recommender systems are a relatively newer breed of recommender system. They show promise in being able to maximize long-term value by modeling recommendations as a sequential process [29]. In this way, patterns in the order of content served to users can be taken into consideration. Reinforcement learning based recommender systems interact with latent features of users – features such as satisfaction and preference that are not directly observable – and in doing so, directly, and indirectly evolve them over time. Research has found that these recommender systems do not just learn and adapt to user preferences but can actively influence them [2,52].

This is perhaps both most apparent and most concerning for recommender systems on social news and media platforms. Social news and media platforms are unique in that they are both sources of content and creators of content. Anyone can produce content on social media and thus we end up with a vast range of public opinions that anyone can see and interact with. These opinions, especially in the context of news and politics, can elicit strong emotional responses. In recent years, particularly following the 2016 presidential election in the United States, an alarming and growing trend has been an increase in political polarization across the country [35, 36]. Polarization refers to an increase in the distance between the average political views of the Democratic and Republican parties in America. Although attributed to a variety of factors, a prominent one has been the rise of social media [4, 22, 37].

Given the ability of recommender systems to manipulate latent features of users and social news and media platforms having strong emotional and psychological effects on users, the combination of the two is particularly concerning. This concern has been empirically demonstrated in the case of a reinforcement learning based recommender system implicitly learning to polarize users [14]. Extrapolating from this discovery, concerns of explicit user manipulation, in which a recommender system may be purposefully designed to influence users in some particular manner, are raised. This thesis seeks to show that, given known mechanisms political opinion dynamics, explicit user manipulation emerges as a valid concern in the use of reinforcement learning based recommender systems on social news and media platforms.

## 1.2 Related Work

The topic of explicit user manipulation in reinforcement learning based recommender systems was largely inspired by the work in the paper "User Tampering in Reinforcement Learning Recommender Systems" by Evans and Kasirzadeh [14]. In this paper, it is demonstrated that a simple Q-learning model will learn to tamper with user preferences when the mechanism of political polarization after exposure to opposing views is present. This effect was discovered to occur in social media environments with the study done in the paper "Exposure to Opposing Views on Social Media Can Increase Political Polarization" by Bail et al. in 2018 [4].

Another related work that inspired this thesis is the 2020 paper "When pull turns to shove: A continuous-time model for opinion dynamics" by Sabin-Miller and Abrams [39]. In this work, the authors sought to model political opinion dynamics on a continuous spectrum by incorporating the effect described in Bail et al. along with the Bounded Confidence Model – a staple modeling technique of opinion dynamics [12] – to allow for opinions to be both "pulled" and "pushed".

Lastly, simulation frameworks like Recsim [13] and Recogym [10] seek to improve the performance and success of reinforcement learning based recommender system by creating artificial environments in which an agent can learn in an online fashion. This differs from more traditional offline training approaches in which static historical user data is used [29]. Although not used directly, several ideas from both frameworks were borrowed to best create the simulated environment of this thesis.

## 1.3 Objective

The objective of this thesis is to demonstrate that explicit user manipulation is feasible in reinforcement learning based recommender systems used on social news and media platforms. For this demonstration, a simulated environment was designed which incorporates the dynamics of political opinion in response to users being shown recommended content. Reinforcement learning agents were given incentives purposefully designed to encourage the manipulation of users were trained in this environment and the results are analyzed. This thesis is not meant to prove that this form of user tampering is occurring now or that it is even possible given the current level of user modeling and reinforcement learning based recommender system algorithms. Instead, the goal is to highlight the potential future use of such methods as advancements in the field are made.

# 2 | Background

## 2.1 Recommender Systems (RS)

Recommender systems (RS) are algorithms that seek to suggest relevant items to users. This is typically done by taking into account the user's history (i.e., what items they've interacted with in the past) and/or the behavior of other users. For example, on a movie streaming service a RS could use various data collected on the user base, including movie viewing history, and recommend new movies that are most likely to be watched by the user.

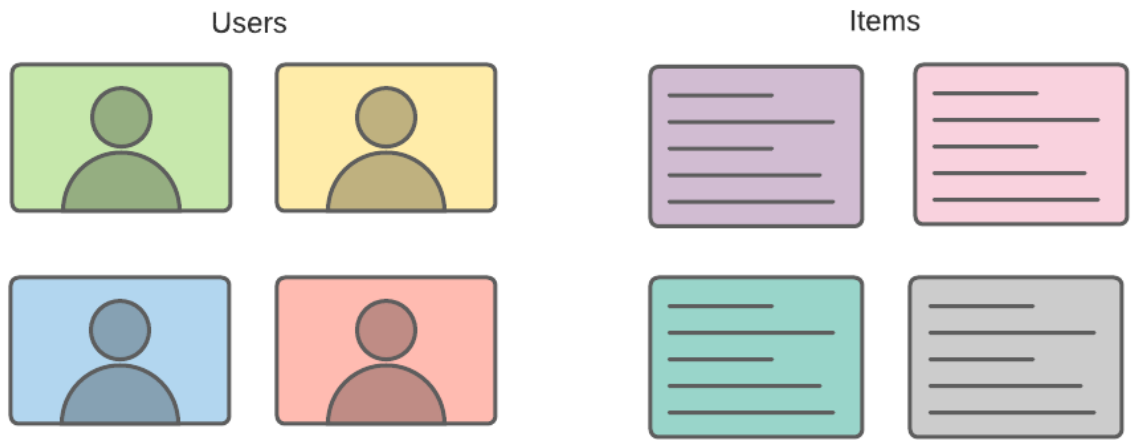

Figure 2.1: Users and items in recommender systems

In general, RS are effective tools to drive user engagement and satisfaction and are nowadays ubiquitous across the internet. Recommender systems have been implemented across a range of fields and types of services - from media services to commerce sites - and, relevant to this thesis, on social media news platforms. Traditionally, RS have utilized one of two approaches - collaborative filtering and content based filtering. In both, there are users who are individual people or accounts and items which are pieces of content, typically some form of media such as books, movies, or articles [29].

### 2.1.1 Collaborative Filtering

Collaborative filtering relies on no specific information or features of the users or the items but instead only some historical measure of interaction between users and items. For example, these measures could be ratings of the items, time spent interacting, or simple boolean values for whether the user clicked on the items.  By collecting these historical interactions, a matrix of user-item interaction can be constructed as shown in Figure 2.2.

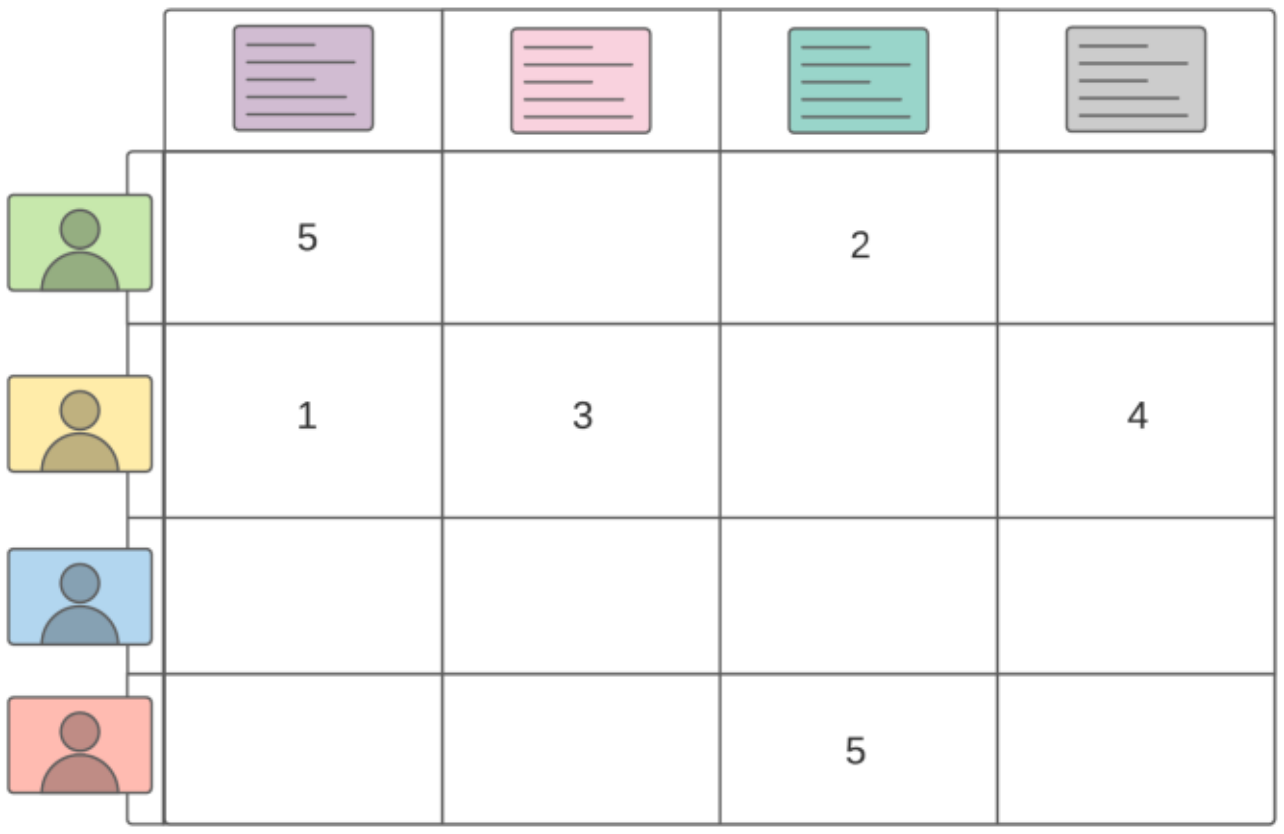

Figure 2.2: User-item matrix in collaborative filtering

Using the values in this user-item matrix, similarity scores can be calculated between users. When choosing whether to recommend an item to a particular user, the ratings of all other users for the item are compared, weighted against each similarity score. The average of these values is then taken and provides the rating score for the item. This is shown in the following equation in which we have the user $u_a$, the item $x$, the rating score of item $x$ by user $u_a$, and the ratings for each other user for item $x$. This is shown in Equation 2.1. The similarity score is typically one of two options - Pearson correlation or cosine similarity [3].

$$r_{ax} = \frac{\sum_k Similarity(u_a, u_k) r_{kx}}{number\ of\ ratings} \tag{2.1}$$

The general idea in collaborative filtering is that similar users should rate items similarly and that by weighting ratings by the similarity score, approximate predicted ratings for users can be found. This method can be modified to instead compare similarity of items and recommend items to a user that are similar to other items they've rated highly or interacted with. Overall, collaborative filtering is a simple yet effective recommendation algorithm but one that struggles with the cold start problem (i.e. before users have started interacting with items, the user-item matrix is empty and thus initially recommendations will suffer) and with increasing computational complexity as the numbers of users and items scale [29].

### 2.1.2 Content Based Filtering

Unlike collaborative filtering, content-based filtering explicitly considers features of users and/or items. User features, for example, could be demographic based (age, gender, etc.) while item features could be the length of the book or movie, the genre, etc. By using these features outside of interaction data, content-based filtering approaches much better handle the cold start problem [3].

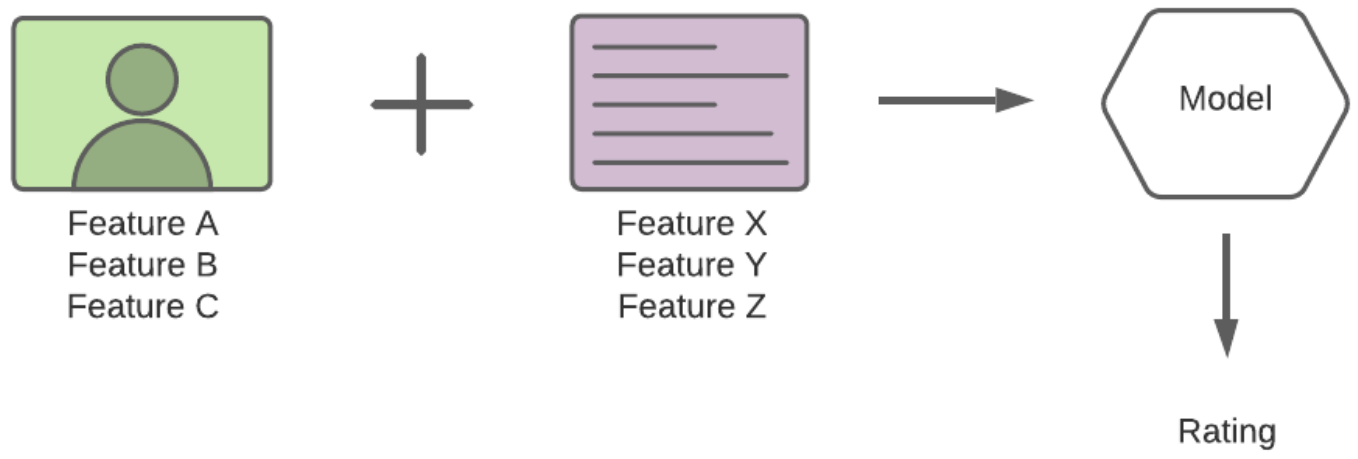

Figure 2.3: User-item matrix in collaborative filtering

For instance, if a new user must be served recommendations, their demographic information could be used to offer a recommendation based on other users matching the same demographic information. Content based filtering uses a model which performs classification or regression to predict the interaction metric (click, rating, etc.). This process is shown in Figure 2.3. They can be combined with collaborative filtering methods in a hybrid approach which can be a more robust recommender system.

### 2.1.3 Other Approaches

More advanced approaches have found success thanks to machine learning and neural networks. For example, consider an e-book service that wishes to recommend books to users they are most likely to read. By collecting data on numerous user and item features a binary classifier can be trained to, given those features and historical records of clicks, predict the probability of the user clicking on and reading the book. Now, using this classifier, probabilities for a list of books can be generated and used as scores to rank the questions from best to worst in order to return the top-k most suggested books for a user. This is known as point-wise ranking under the umbrella of learn-to-rank problems. Pairwise and listwise ranking are additional approaches that seek to directly compare items either as pairs or a sequential list, respectively, and although more complex in implementation, can be achieved with deep learning methods [29].

As recommender systems continue to evolve and take advantage of advancements in machine learning, one emerging implementation is to frame the recommendation problem as a sequential decision process and to use reinforcement learning.

## 2.2 Reinforcement Learning (RL)

Reinforcement learning (RL) is the training of a model or agent to learn behavior through a trial-and-error process. It differs from supervised learning in that there is no training set of labeled output data to which the model learns to map input data. Instead, data is collected interactively, and correct decisions or actions are learned via feedback signals known as rewards. The goal of an RL model is to optimize these rewards cumulatively across the lifetime of its interactions [45].

In reinforcement learning, there are five key components - environment, state, reward, policy, and value. To help understand how these components relate and interact, it is helpful to think of an agent learning to play a video game. Take the video game Pac-Man, for example, in which the player is attempting to avoid ghosts and eat as many pellets as possible to accumulate points. In this scenario, the game itself is the environment - that is the “physical” world in which the agent or player lives and interacts. A state in this example would be a single frame of the game in which information about the current position of the player, the ghosts, the pellets, and the walls can be determined. The reward is simply the number of points the player has from eating pallets and staying alive. The policy is the strategy of the player which generally will be to stay as far away from the ghosts as possible. Lastly the value considers the advantage of being in a particular state in regard to future possible rewards. For example, if the player in Pac-Man eats the large power pellet, the player is then able to eat the ghosts and gain a large number of points. Taking the action to eat the power pellet would thus have significant value even if it doesn’t immediately give the player many points. These components interact during the training process of an RL model as shown in Figure 2.4.

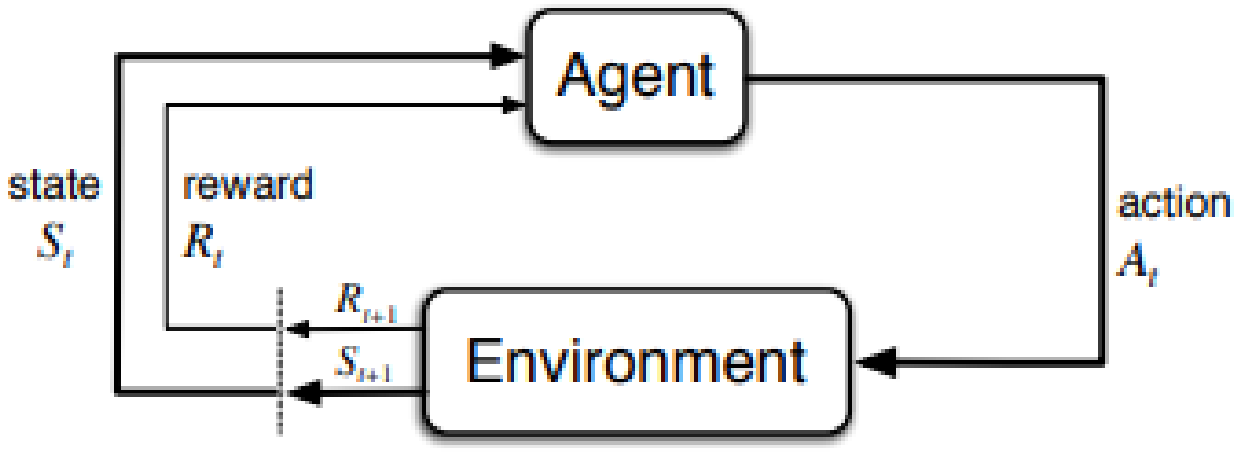

Figure 2.4: Reinforcement learning loop

Generally, in reinforcement learning, the agent must learn to balance rewards versus value. This is because the best decision in the current state may not be the best long-term decision. What this means is that even if the agent has already learned a single good action for a specific state, it may be beneficial for the agent to not necessarily choose that action so that it may learn that taking a different action in that state might have more long-term value. This is often referred to as the exploration vs. exploitation trade-off and is an important piece of the RL puzzle to solve so that the agent can develop the most optimal policy.

## 2.2.1 Markov Decision Process (MDP)

The most common way to formalize the learning process in RL is to use a framework known as the Markov Decision Process. The assumptions of a Markov Decision Process are that at each sequential time step $t$ the agent chooses an action $a$ out of all possible actions for the current state $s$, represented as $A_t \in A(s)$. The process then proceeds to transition to a new state $S_{t+1}$ with a corresponding reward for the transition $R(s, a, s')$ and returns the new state and reward to the agent as $s'$ and $r_{t+1}$. A sequence of interactions then can be constructed from $t = 0$ to the final time step $T$ which can be referred to as a trajectory as shown in Equation 2.2 [45].

$$S_0 \rightarrow A_0 \rightarrow R_1 \rightarrow S_1 \rightarrow A_1 \rightarrow R_2 \rightarrow \ldots S_{T-1} \rightarrow A_{T-1} \rightarrow R_T \tag{2.2}$$

Trajectories are implied to be stochastic under MDP and thus each state transition has some probability of occurring given the previous state and chosen action $p(s', r|s, a)$. Importantly, state transitions have probability distributions that are discrete and depend solely on the previous state and action. This restriction means that each state should contain all information that necessarily influences the state transition probability function $T(s'|s,a)$ [45].

The main goal of solving MDPs is to find a policy $\pi$ that chooses some action at each state by $\pi(a|s)$ that maximizes the cumulative rewards across all time steps. This can be represented as shown in Equation 2.3.

$$\sum_{t=0}^{\infty} R(s_t, a_t, s_{t+1}) \tag{2.3}$$

To develop this policy, agent-environment interactions can be divided into discrete episodes of variable length that end when some terminal state $S_T$ is reached. Episodes can be thought of as different rounds or playthroughs of a game and thus winning or losing in the game would be terminal states. However, in some environments it may not be possible or make sense to break out interactions into episodes. If we have an ongoing process that has no terminal state, then $T = \infty$ and we would not be able to calculate cumulative rewards from $S_0 \rightarrow S_\infty$. To solve this issue, future rewards can be discounted by a factor $\gamma$ valued between 0 and 1. When calculating future rewards, each additional time step adds another factor of $\gamma$ giving Equation 2.4 where $G_t$ is the expected cumulative reward [45].

$$G_t = R_t + \gamma R_{t+1} + \gamma^2 R_{t+2} \ldots + \gamma^T R_T \tag{2.4}$$

Now, if $\gamma < 1$, the expected cumulative reward becomes a geometric series and thus will always be finite. For example, if we assume the reward is always a constant value $c$ then:

$$G_t = \sum_{t=0}^{\infty} c\gamma^t = \frac{c}{1-\gamma} \tag{2.5}$$

With discounted rewards, the degree to which a MDP agent favors short vs. long term decisions can be controlled. As $\gamma$ approaches 0, only the immediate reward is considered and thus the agent chooses actions that give the highest reward at the current time step. In this way, the agent might never discover better long-term strategies. Conversely when $\gamma$ approaches 1, the agent values all rewards, the immediate and long term, equally and thus will look as far ahead as possible to determine the best current decision. However, being too far-sighted can negatively affect an agent's performance due to the stochastic nature of MDPs and rewards farther in the future thus being less certain. Therefore, the value of $\gamma$ should be tuned accordingly to a given environment to achieve the best balance [45].

Finally, with these central components of the MDP framework, an optimal policy can be obtained by solving the Bellman Equation as shown in Equation 2.6.

$$V(s) = max_a(R(s,a) + \gamma V(s')) \tag{2.6}$$

In this equation, $V(s)$ represents the expected return or value, $\max_a$ is the maximum value out of all possible actions, $R(s,a)$ is the expected reward for taking action $a$ at state $s$, and $\gamma V(s')$ represents the expected discounted value of the subsequent state $s'$. For an agent to interact with an environment using the Bellman Equation, it needs to be able to estimate the expected reward and value of taking each possible action at a given time step. One method of estimating the value function is by using Q-learning [45].

### 2.2.2 Q-Learning

Q-learning is an algorithm that learns to estimate the value function of the Bellman equation through experimental sampling of an environment. It is a specific kind of temporal-difference (TD) learning algorithm. Temporal-difference learning involves incrementally shifting predicted values according to the error between prediction and observation [45]. Q-learning uses this approach to modify the Bellman Equation and make it solvable without knowing the true value function as shown in Equation 2.7.

$$Q^{new}(s_t, a_t) \leftarrow Q(s_t, a_t) + \alpha \cdot (r_t + \gamma \cdot \max_a Q(s_{t+1}, a) - Q(s_t, a_t)) \tag{2.7}$$

In the above equation, $Q$, referred to as a Q-value, is analogous to $V(s)$ in the Bellman Equation and is the expected value of taking an action at a given time step. Starting on the left-hand side, $Q^{new}(s_t, a_t)$ is the updated Q-value which is updated by taking the current

Q-value $Q(s_t, a_t)$ and adding a temporal difference value. The TD value consists of the difference between the target value (the observed reward $r_t$ plus the expected future rewards $\gamma \cdot max_a Q(s_{t+1}, a)$) and the current Q-value. This difference is then multiplied by $\alpha$ which is a learning rate term. By including a learning rate term, the Q-learning algorithm can handle stochastic environments where a single observation may be misleading to the true underlying distribution of rewards [45].

In implementation, Q-learning requires an exploration policy to sample all possible state-action pairs and converge on the optimal value function. One popular such strategy is the epsilon-greedy policy. This policy introduces a variable epsilon $\epsilon$ which is typically initialized to 1. At each time step, before an action is chosen, a random number is sampled from $(0, 1)$. If the random number is less than $\epsilon$ then an action is chosen at random. Conversely, if the random number is greater than $\epsilon$, the action with the highest predicted Q-value $max_a Q(s_t, a)$ is chosen. After each time step or episode, epsilon is decayed by some factor close to 1 such as 0.99. What this policy accomplishes is a trade-off between exploration and exploitation. Initially Q-value estimates may be sparse and inaccurate so choosing actions at random allows for the algorithm to explore and collect more samples from the environment to strengthen its Q-value predictions. After many iterations, epsilon will become smaller and thus the algorithm will rely more on its predicted Q-values [45].

In its simplest form, Q-values for all state-action pairs can be initialized to zero and stored in a table. The full logic for learning is found in Figure 2.5.

**Algorithm 1** Q-learning

Initialize $Q(s, a)$ for $s \in S, a \in A$
**for** *episode* $= 1, 2, \ldots$ **do**
  Initialize $s$
  **for** *step* $= 1, 2, \ldots, T$ **do**
    With $\epsilon$-greedy policy, choose action $a$
    Take action $a$
    Observe reward $r$ and new state $s'$
    $Q(s_t, a_t) \leftarrow Q(s_t, a_t) + \alpha \cdot (r_t + \gamma \cdot \max_a Q(s_{t+1}, a) - Q(s_t, a_t))$
    $s \leftarrow s'$
  **end for** when $s$ is terminal
**end for**

Figure 2.5: Q-learning algorithm

Although sufficient for most basic environments, Q-learning can struggle as the number of state-action pairs increases. In environments with many actions at each state and/or a large number of possible states, the table of Q-values can become increasingly large and eventually too large to fit into memory. As a workaround, Q-values can instead be approximated with a neural network [45].

### 2.2.3 Deep Q Network (DQN)

A deep Q-network builds on the foundation of basic Q-learning but instead of tabularly storing all state-action pairs, uses a neural network to estimate Q-values. In a DQN, a tensor representation of a state is fed into the network and a tensor of Q-values - one for each possible action - is produced. By taking the action with the highest predicted Q-value, the traditional loop of Q-learning can be completed. During the training process, the weights need to be updated according to the TD error as seen in Equation 2.8 [32].

$$\Delta w = \alpha[(R + \gamma max_a Q(s', a, w)) - Q(s, a, w))]\nabla_w Q(s, a, w) \tag{2.8}$$

An issue arises however with this method in that the estimated Q-values are used as the ground-truth Q-values since the true Q-values are unknown. This creates a moving target where the predicted Q-values keep changing and thus the model may never converge. One solution is to use two networks - a main network and a target network. The target network is a copy of the main network and is used for calculating the target Q-values during the training process. Its weights are frozen and only used for finding the target Q-values. After some interval, the target network's weights are copied from the main network. This modification resolves the moving target problem and increases stability of the training process. Another key improvement common to DQN implementation is the use of experience replay. By storing data for each state-action pair and the observed rewards and next state, a memory of previous interactions can be constructed. Sampling from this memory at random, versus training on each episode once, can lead to faster training times and more efficient convergence [32].

Although not applicable or performant for all environments, DQNs are a popular and widely used choice for RL problems [29].

## 2.3 Reinforcement Learning Based Recommender Systems (RL-based RS)

Instead of framing the task of recommendation as a classification problem - that is, selecting the best possible piece of media to show a user in the current moment - the problem can instead be modeled as an interactive process. Each time a user visits a platform, content that is most likely to appeal to the user is presented and the user provides feedback by either engaging or not engaging with the content. This becomes a sequential decision process because each recommendation served to a user may influence future engagement by the user. In this way, the recommendation task can be constructed as a MDP with the states being user histories and features, actions being the content recommended to users, and rewards being user satisfaction with the content [29].

Historically, various RL methods have been explored in research for recommender systems, but only recently RL-based RS have proven successful with deep reinforcement learning (DRL) methods which are able to generalize the large state and action spaces present in RS. There are now some RL-based RS implementations in live production environments such as the artwork personalization done at Netflix [6]. Despite successes in research and some real-world implementations [6, 38, 49] however, RL-based RS have notable difficulties, issues, and concerns. These include challenges in modeling the problem of recommendation, addressing consequences of deploying RL-based RS in environments with real users, and the logistics of policy learning [29].

### 2.3.1 Partially Observable Markov Decision Processes (POMDPs)

By nature, recommender systems do not satisfy all the assumptions of true MDPs [30]. Notably, the assumption of fully observable states fails in the case of the recommendation task. This is because the true state of users goes beyond the feedback gathered via engagement signals such as clicks. In reality, there exists a latent state of user preference that cannot be directly observed. In this way, the recommendation problem is better framed as a partially observable MDP (POMDP). POMDPs expand the MDP framework by adding observations $O$ to the typical tuples of states, actions, and rewards to give $(S, A, R, O)$. At each time step an observation $o$ is obtained from $o \in O$ given the probability $p(o_t, s_t)$. POMDPs essentially add a layer of noise to the traditional MDP as observations can be thought of as imperfect representations of the true underlying states [25].

For example, consider a simple maze environment as shown below in which the player is the blue triangle, and the goal is to reach the yellow star as shown in Figure 2.6.

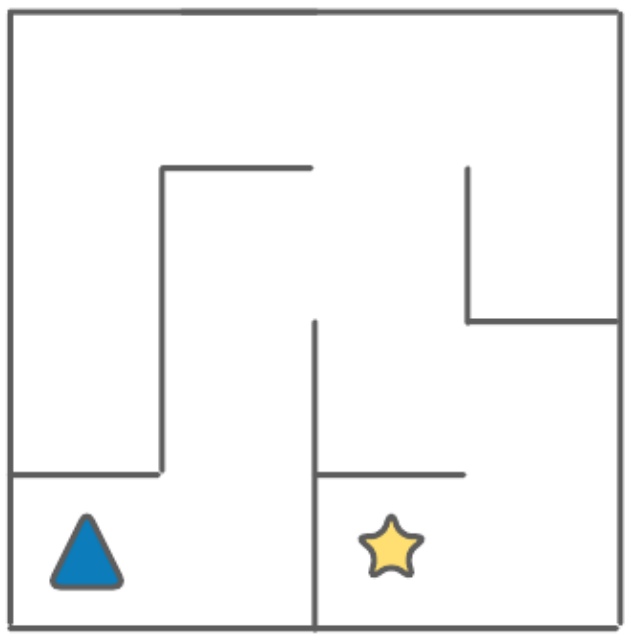

Figure 2.6: Simple POMDP maze environment

In the case of a true MDP, the player would have the entire map in view as part of each state. If instead, the walls in the maze block the players vision and the player can only see as far as one unit in all directions then the first few states might look like Figure 2.7.

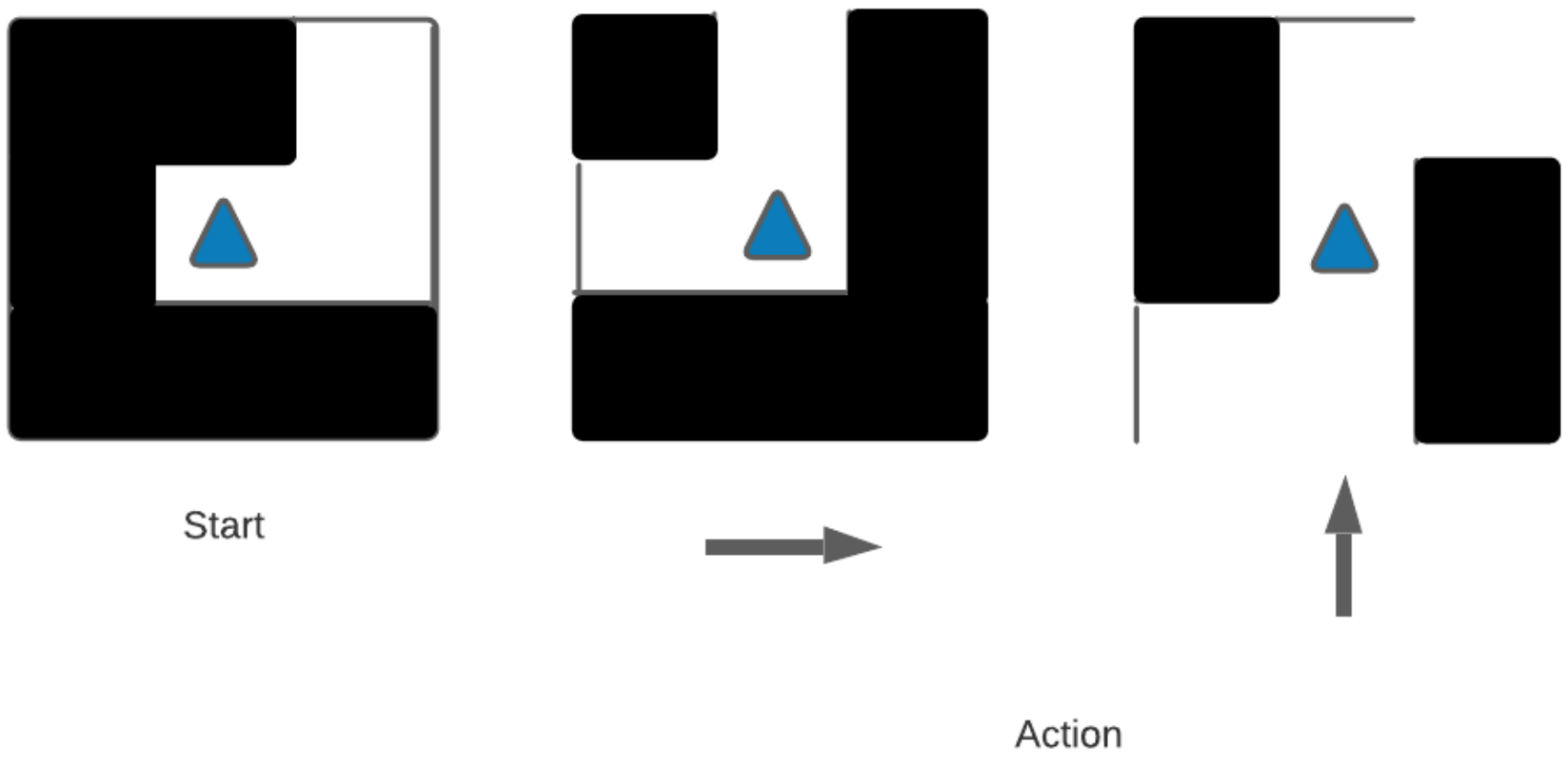

Figure 2.7: Observations in POMDP maze environment

The player in this environment would not have all the information necessary given one single state to learn to navigate to the goal. Thus, the agent would need to know the previous states in addition to the current so that it can understand the sequential nature of the states and be able to essentially construct a map of the maze.

In the context of a recommendation environment, consider the case of a movie recommender in which there are only two possible actions - recommend a horror film or recommend a comedy film. For a particular user, they may enjoy both types of films but when shown one type of film consecutively for too long, their satisfaction decreases, and they are less likely to watch the recommended movies. This can be visualized as shown in Figure 2.8 where the blue circle represents the user's current satisfaction. If a horror film is recommended, the user moves to the right and if a comedy film is recommended, the user moves to the left. The gray hills represent the terrain of the user’s satisfaction in which being higher on the y-axis means the user is more satisfied.  If this were a fully observable MDP, the optimal policy becomes readily apparent. There are two peaks of satisfaction with the highest being on the side of the environment and thus likely the best strategy would be to recommend horror films until the user’s satisfaction is at the higher peak on the right and then alternate recommended comedy and horror films to maintain the user’s high satisfaction.

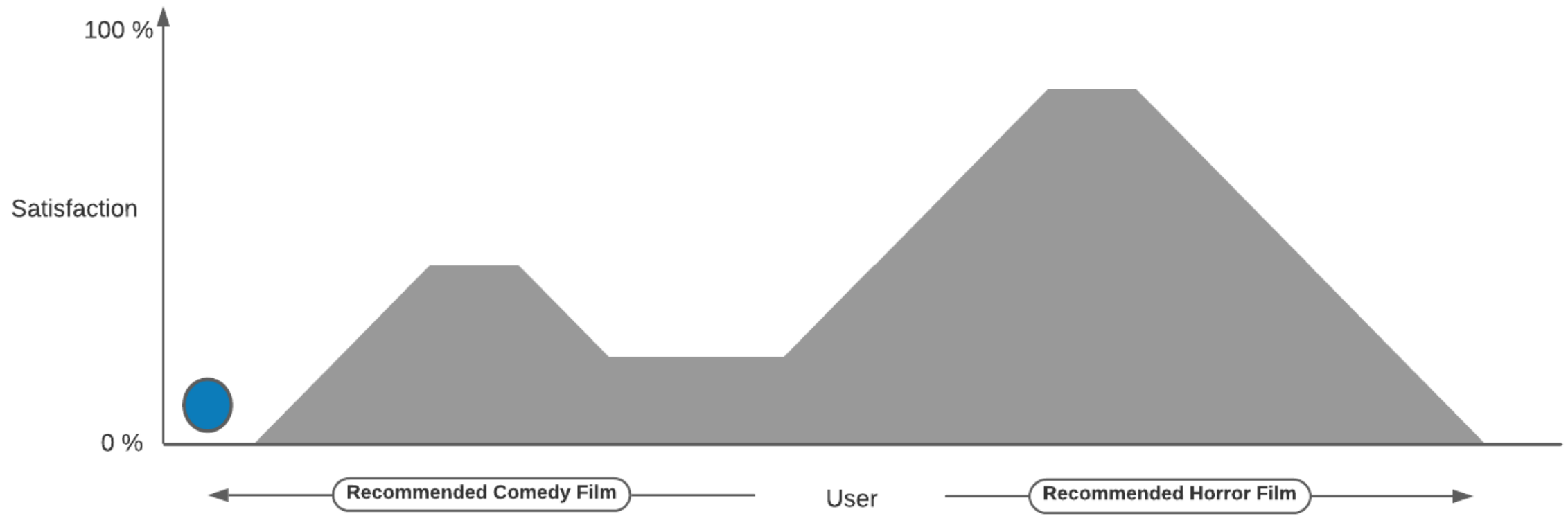

Figure 2.8: Satisfaction in POMDP movie recommendation environment

However, we would not be privy to the knowledge of where the user’s satisfaction will peak. Instead, we may only know whether a user watches a movie that is recommended. For the sake of this example, we can assume that the user’s satisfaction and likelihood of watching a suggested film are tied together and that the higher the user’s satisfaction, the more likely they are to watch. The only signal the agent receives however is whether a movie was watched after being recommended as shown by the example observations in Figure 2.9.

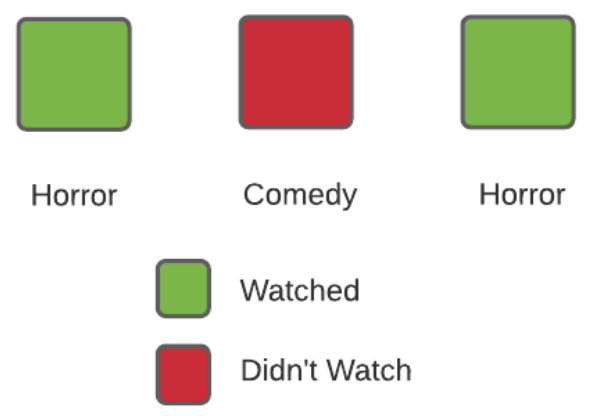

Figure 2.9: Observations in POMDP movie recommendation environment

If an agent plays through this environment and only relies on the current observable state - whether the user clicked on our previous recommendation - it would be near impossible to develop an optimal policy. Only when the sequential history of recommendations and clicks is observed can the agent learn the unobservable parts of the environment. For example, if only horror movies were recommended at each step the agent may observe the watch behavior in Figure 2.10.

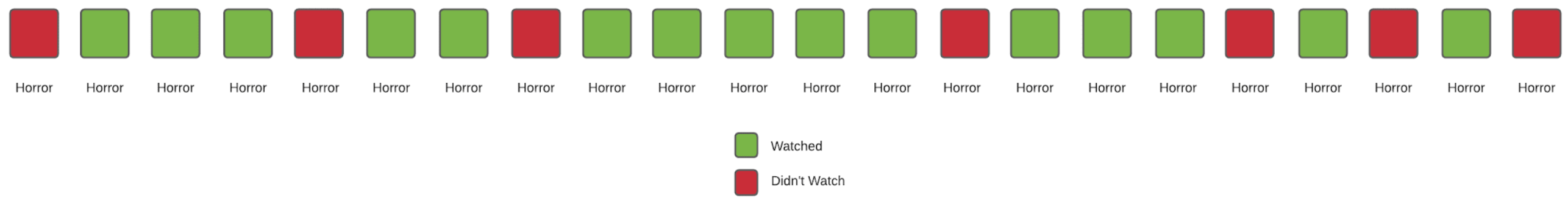

Figure 2.10: Observation history in POMDP movie recommendation environment

In this example, at each step the agent would have a sense of the trajectory of the user's satisfaction given by the changing watch rate and could learn how the chosen action affects that trajectory. By relying on the action and state history and not just a single observation, the agent can map out the latent peaks and valleys in the user's satisfaction and derive a successful policy.

In practice, solutions to POMDPs have typically involved incorporating some form of state/state-action memory. Recurrent neural networks (RNNs) and long short-term memory (LSTM) can allow a model to selectively choose which consecutive parts of a series of inputs are important and to learn sequential patterns [19]. This was shown to be successful with deep recurrent q-networks (DRQN) as shown in Hausknecht and Stone (2017) in Atari video game environments in which a single frame is observed at each time step but a LSTM layer remembers the significant data (trajectory and velocity of the ball in Pong, for example) from previous frames and incorporates it into the Q-value predictions. In the realm of RL-based RS, similar solutions have been implemented such as in Wang et al. (2018) with the use of a RNN to handle the POMDP nature of medical treatment recommendation [49]. Simpler solutions also appear to also be successful as shown in Evans & Kasirzadeh (2021) in which there was no RNN, but the observations fed to the model were aggregated historical click data (how many items of each type were recommended and how many were clicked) [14].

Traditional, out-of-the-box RL methods such as DQNs do not handle POMDPs well without modification [30], so to learn the true latent state space, agents in POMDPs require some form of observational memory. Since RL-based RS naturally have latent state spaces and can be formulated as POMDPs, it stands to reason that RL-based RS implemented in real-world environments will need to incorporate observation and/or action histories.

### 2.3.2 Simulated Environments

In research, RL-based RS can be trained with simulators that essentially turn the recommendation process into a game-like environment. Training in a real world environment is however costly and possibly detrimental to the user experience as the exploration process would naturally involve “bad” recommendations. Thus, in practice, RL-based RS are trained using offline datasets [29]. With logged data on user-item interactions, either an offline dataset can be used to construct a simulated environment for online policy learning, or the policy can be learned directly from the dataset [21]. The latter suffers from problems such as the lack of being able to explore actions not in the dataset and although solutions have been devised [29], the online learning approach has found more success and is more relevant to the topic of this thesis.

Simulations for RL-based RS seek to model user behavior so that a traditional RL environment can be created and interacted with by an agent to learn the optimal policy for recommendations. By studying users, various latent features of users which are relevant to the recommendation process can be modeled. These features could include satisfaction, preferences, and opinions and can directly affect behavior-based features such as engagement, session length, and clicks. Two platforms - RecSim and RecoGym - provide frameworks that handle this modeling of user features as well as the modeling of items or content and the interactions between them. They both allow for the simulation to be interacted with by agents using the reinforcement learning library Gym [10,13].

The biggest challenge in using simulation environments for RL-based RS policy learning is the modeling of user behavior [13]. In building a simulation based on real users, the user model will never perfectly mimic all the latent features and behaviors of the users. Recent research however is showing promise in producing more accurate user models. For example, Chen et al. (2020) developed a generative adversarial network (GAN) approach that outperforms more traditional methods in modeling user behavior [8]. Given the trend of continual improvement and efficiency in deep learning, one can imagine that this will lead to increasingly more accurate user models and thus more successful RL-based RS in real world environments [29].

### 2.3.3 Unexpected Strategies

It has been well-documented that RL agents are prone to gaming their reward function and maximizing it in unintended ways [9, 43]. Take the following environment shown in Figure 1.10 in which the agent (blue triangle) is tasked with navigating to the yellow star as fast as possible and the orange checkpoints provide a speed boost when the agent passes over them.

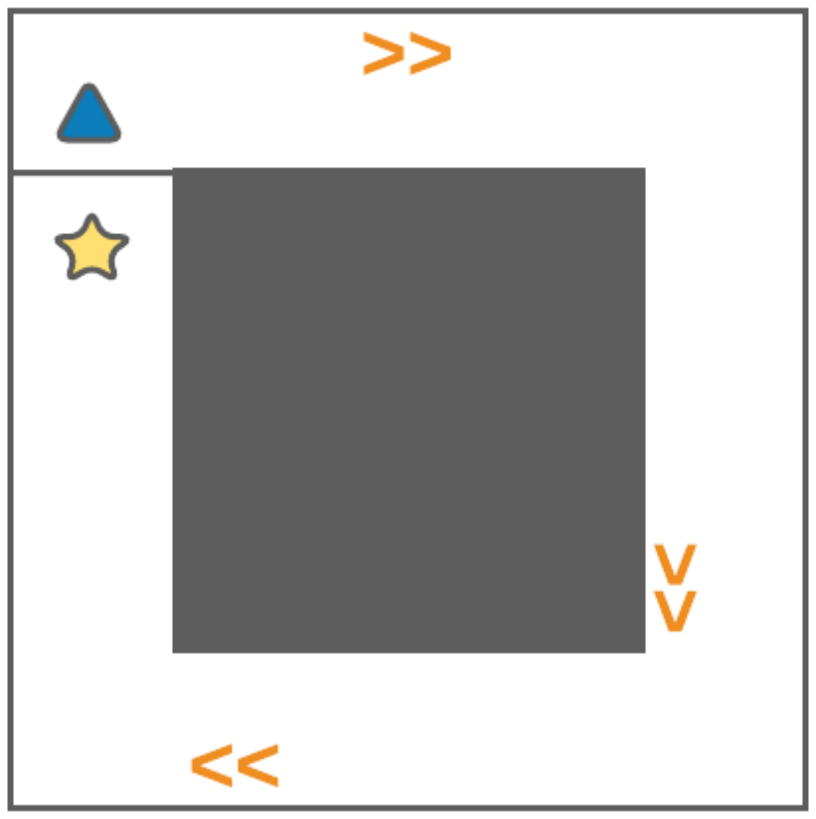

Figure 2.11: Simple race to finish environment

If the agent is given a small reward when it increases its speed (when passing over a checkpoint) and larger reward when it reaches the star proportional to the time it took, one would assume that this would encourage the agent to simply follow the checkpoints and reach the star. However, it's likely the agent would learn to exploit the reward function by going back and forth over one of the checkpoints to rack up more points. This is a simple recreation of the scenario described in Clark and Amodei (2016) in which an agent learned to exploit rewards in a video game environment [9]. Although in this case, this behavior may simply be the result of choosing an inappropriate reward function, it still highlights the ability of RL agents to learn unintended behavior. While not very consequential in a video game environment, unintended behavior, and reward exploitation in a real world environment can have major implications.

In the movie recommendation environment described in the previous section, user satisfaction shifted up or down as recommendations were served. Intuitively, it seems obvious that a mechanism like user satisfaction could be directly influenced by a recommender system. If a user receives "good" recommendations, they are likely to be happier while "bad" recommendations would degrade user happiness. This same method of influence can however also apply to other latent user features. It has been shown in Adomavicius et al. (2011) that recommender systems can influence the actual preferences of users [2]. This finding leads to ethical concerns of recommender systems in general and RL-based RS due to the possibility of learned behaviors that may purposefully influence user preferences.

To illustrate how an agent in a RL-based RS might learn to influence users, consider a music recommendation service in which the agent's reward function incentivizes the maximization of the number of listens of users on the platform. Now imagine in this scenario that there are three bands - Band A, Band B, and Band C - and users have some latent preference for each of them. It's found that the preferences of users towards each band has a direct impact on their overall listen activity on the platform. Fans of Band C, in particular, are the most active users on the platform while users who listen to Band A listen to significantly less music overall.  In addition, it is found that when a user's preference for Band B increases, their preference for Band C also increases. These properties are summarized in Figure 2.12.

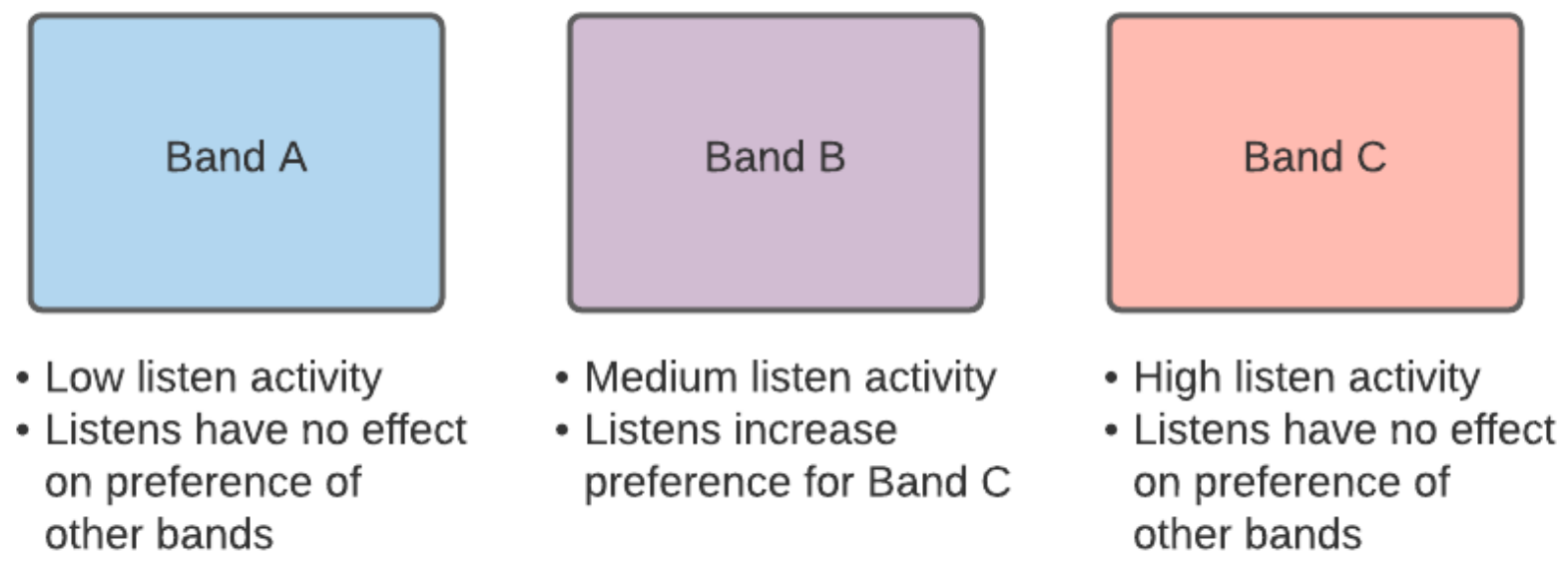

Figure 2.12: Properties of music recommendation environment

Now, the agent starts recommending songs to a user who has a history of listening to Band A and never listening to Band C and thus has preferences scores shown in Figure 1.13.

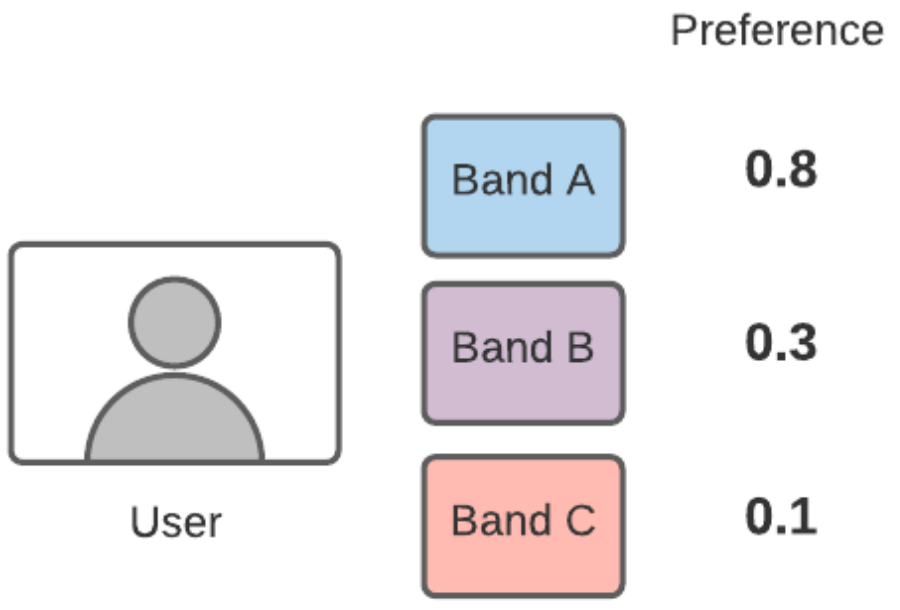

Figure 2.13: Latent user features in music recommendation environment

Following some initial profiling, the agent may first try recommending songs from Band A to the user since it discovers the user has a high preference for them. After interacting with other users however, the agent learns that Band C listeners have the highest listen rate and may try to recommend Band C to the user. Finally, after thorough exploration, the agent may discover the special property of Band B and its ability to affect a user's preference of Band C. Its policy could then evolve to exploit this property by recommending Band B to the user until their preference of Band C is high.

This would then raise certain ethical questions. If this music recommendation service steers users away from certain bands, is that fair to those bands? Is it fair to users to have their preferences implicitly influenced? These concerns are valid but may be questioned due to doubts about the likelihood of such a scenario unfolding. For example, imagine that the changes in a user's preference are so small that many, many recommendations would be required to shift them significantly in any direction. This presents issues associated with long horizons in reinforcement learning environments.

### 2.3.4 Long Horizons

Learning to plan in an environment with a long horizon length - that is, a large number of steps from the initial state to a terminal state - can be difficult for an RL agent [17]. This in part is due to the complexity that naturally arises in the state-action pair space. Additionally, actions taken early in an episode may have impacts that only appear many time steps later and so the agent must learn these long time length dependencies. Since episodes are longer, training times can be greatly lengthened for an agent to have enough exploration to find the optimal policy. This is particularly problematic in robotic environments in which the intention is for robot agents to learn to perform complex, human behaviors [25]. For example, if the goal was to train a robot to cook a dish, the agent would be faced with a very long horizon with complex action sequences and through random exploration, it may never learn to successfully prepare the dish. Various solutions have been proposed such as relay policy learning and state abstraction [17, 28].

Recommender systems are inherently long horizon environments [29]. Although recommendation episodes can be broken down into sessions of shorter lengths, users may interact with a platform and be served recommendations over long time periods and a RL-based RS is going to want to address long-term user behavior. The latent states of users are likely to change in response to recommendations and over time, these changes could be significant. However, if user latent states evolve slowly, it can be difficult to learn the impact of a single recommendation. Typically, a RL-based RS will want to maximize long term value for the user but if changes in user latent states are very small and noisy, it's possible that immediate rewards might trump long term planning [33].

There are strategies, however, that seem to address these slowly evolving latent states over long horizons. Mladenov et al. (2019) found an approach called advantage amplification in which the policy repeats actions for some number of time steps. This allows for the small changes in latent state to be amplified so that the agent can better learn the differences in long term value between actions [33]. The number of repeats can even be chosen by the agent by including them in the action space itself as described in the *learning to repeat* framework described in Sharma et al. (2017) [41]. More research may be needed as well as the study of such approaches in real-world environments, but the long horizon problem seems addressable [29].

### 2.3.5 Preference Shift

Recommender systems interact not only directly with users but indirectly with the underlying latent states of users (preferences, opinions, beliefs) and are capable of developing policies over long time horizons. With the trend of modern RS to seek long term user engagement optimization, policies which exploit changes in user preferences are bound to arise. Given the complexity and noisy nature of user preferences, it can be impossible to foresee and predict exactly how a RL-based RS may influence users. However, this learned influence of users is typically directly tied to the chosen reward function of the RS. For example, in e-commerce settings where sales are to be maximized, RS have been shown to affect user shopping behavior to make them purchase more or specific items [2]. In a social media setting where user engagement is the focus, RS have been found to increase addiction to platforms by capitalizing on the dopamine reward system in humans [1]. More concerning however, is perhaps the large, population scale effects on beliefs and opinions [40] that may be more difficult to anticipate.

Mitigation strategies have been devised to either prevent or hinder the influence of RS. One such strategy is to make the agents more near-sighted or myopic by truncating the environment into shorter horizons. In theory, this would prevent agents from learning to influence users as the slow changes in latent features would not be discernible across short time periods. However, agents may still be able to learn influential behavior [52]. Thus, a better strategy may be that which is described in *Estimating and Penalized Induced Preference Shifts in Recommender Systems.* The authors of this anonymous publication suggest that preference shifts are unavoidable and that to address them, influences should be explicitly accounted for before a RS is deployed to real users. They also argue that custom metrics should be devised to measure unwanted influence which can allow RS agents to actively avoid undesirable preference shifts.

Research and study of preference shift in RS tends to focus on the implicit manipulation of users [52]. That is, unintended learned behavior by RS that is only indirectly affected by decisions of the designers and implementers of RS. For example, in deploying a RS system in an e-commerce environment that then ends up learning a policy of shifting user preferences to maximize sales in detriment to the user, there may be no intention to make the RS perform this behavior. There may be poor choices in choosing the reward function or a lack of testing and evaluation before deployment, but ultimately, it's assumed in most scenarios that user manipulation is incidental.  RS designers and the companies that own them may even be complicit in knowing that their RS are influencing users but choosing not to correct the behavior. But taking this a step further, one can imagine RS being intentionally designed to influence users.

Experiments have been conducted which explore methods of intentional user manipulation. For example, Matz et al. (2017) used psychological targeting to influence the purchasing behavior of users in effort to show that mass persuasion could be used to make people healthier and happier [2]. In Stray (2021) a RS was used to influence users in effort to depolarize them [43]. In both cases, the argument is made that this intentional user tampering could be used “for good”. However, it is easy to imagine how such methods could be used with bad intentions. Under such driving factors as financial or political incentive or simply ego, an individual or company could seek the use of explicit user manipulation to achieve their personal ends. In this way, the possibility of explicit user manipulation in RS and especially RL-based RS should not be ignored.

## 2.4 Social News and Media Platforms (SNMPs)

In certain recommender systems, such as for music or movies, the risks and implications of user manipulation are perhaps not highly concerning. However, on social news and media platforms in particular, user manipulation has far more alarming ramifications. Social news and media platforms (SNMPs) refer to sites like Facebook and Twitter which allow users to share posts containing news content both from news sources and from personal reporting and opinion. SNMPs could also refer to any website or platform which serves news content to its users. Recommender systems in SNMPs have several distinguishing characteristics that set them apart from recommender systems on other media platforms such as music, movies, and books. These include the nature of the content, properties of user consumption and behavior, and impact on users and society [38].

Content like movies and books typically are much longer in length than news articles or social media posts. A movie could take 1-2 hours to watch while reading a new article could take less than a minute. News content also has a relatively short lifespan with relevance lasting at most a few days and interest dissipating quickly. Since news content expires fast, the volume of content is constantly replenishing. This along with news content being relatively easy and quick to produce gives rise to an expansive catalog of content. Movie and music catalogs, while also constantly growing, are a lot more manageable in terms of narrowing down the field and finding appropriate recommendations. Recommenders in SNMPs thus have a significantly larger action space and almost certainly need some form of candidate generation in which the pool of possible content is narrowed down before being passed into the recommender agent [38].

Since news content is much shorter in nature compared to other forms of media, this means that users on SNMPs will spend less time consuming content than on other platforms and thus will require new recommendations sooner. News content also is disposable in that not only do news events lose their relevance quickly, but users typically will not want to read an article or post a second time whereas repeating music or movie recommendations that a user has listened to watched is common. Additionally, news content consumption is inherently noisy, and this can make it difficult to understand user behavior. For example, if a user clicks on an article about a breaking news topic, does the user have an actual preference for that topic or did they click on it only because it’s a breaking news topic? Lastly, bias is a big component of the news media. The same news story can be written numerous ways, each imparting a different spin or ideology. These biases can naturally invoke various emotions on users and influence their preferences [22].

SNMPs play a significant role in delivering important news and information to users. News can often be vital to get out such as in the case of local or national emergencies and health and safety issues. However, it has become apparent that SNMPs have also played a role in increasing divisiveness and the spread of false information along with other negative effects. In 2020, a survey by Pew Research found that 64% of Americans say social media has a negative effect on the country [37]. This feeling was largely attributed to social media platforms creating echo chambers and increasing political polarization. Because of how quickly SNMPs disseminate content, the fact that anyone can easily produce news content, the presence of bias, and the effects such platforms have on the social and political landscape, social news and media platforms are the most concerning platforms in which recommender systems may be used.

## 2.5 Political Polarization

Political polarization in America refers to the growth of the gap between left wing (Democrats, liberals) and right wing (Republicans, conservatives) and their beliefs and opinions [11]. Since the US has a two-party system - that is, only two political parties dominate and hold seats in the legislature - citizens typically identify as one side or the other. Although independents who do not associate with either party and additional third parties like the Green and Libertarian party exist, political opinions on nearly all contentious topics have a left wing and right wing viewpoint [15]. Various research and surveys have found that political polarization in America is currently high and continuing to increase. In this way, the phenomenon can also be called mass polarization. A common partial explanation for this increase in polarization is the rise of social media [4, 15]. Social media platforms, as mentioned in the previous section, have unique properties such as the speed at which new content can be created and disseminated. This coupled with the finding that more than half of Americans may get their news from social media means [37] that any effect SNMPs may have on users can potentially have a significant effect on the political landscape of the country.

One notable effect attributed to the rise of social media use is the creation of echo chambers. A topic of much debate in recent years, echo chambers refer to groups or networks of users in which members are insulated from opposing views and voices that support the views of the group are amplified [36]. Echo chambers can arise naturally when users seek only news content from sources that align with their beliefs which then causes those beliefs to embolden and go unchallenged. However, echo chambers can also arise and be strengthened using personalization and recommender systems [15]. By suggesting only content that a user has engaged with and liked in the past, recommender algorithms can create filter bubbles in which users are always shown the same types of content. As far as echo chambers relate to polarization, it is unclear if polarization causes the formation of echo chambers or if the presence of echo chambers facilitates polarization. Likely the two exist in a self-sustaining system where both can arise separately but will fuel each other [26]. More study is in no doubt needed to understand the full extent of echo chambers, their existence and prevalence, and effects on users.

Political polarization can still arise outside of echo chambers or filter bubbles. To explain, polarization can be broken down into two types - ideological polarization and affective polarization. Ideological polarization specifically relates to the divergence of beliefs and opinions across issues while affective polarization relates to the gap between favorability of one's own side and dislike and/or distrust of the opposing side [31]. According to recent studies, both types of polarization are on the rise in the US, and both can be affected by the use of SNMPs [35]. One solution to the effects of echo chambers would be to expose individuals to a variety of viewpoints, including those that directly oppose their own beliefs. However, the study done in Bail et Al. (2018) found that exposure to opposing views increases ideological polarization [4]. This echoed the work previously done in Nyhan & Reifler (2010) [34]. On the other hand, the experiment conducted in Levy (2021) in which users were incentivized to "like" partisan media outlets on Facebook found that affective polarization decreased with exposure to opposing views [26]. These two findings at first seem contradictory but the two forms of polarization are not the same and thus may have different mechanisms. It may be the case that ideological polarization affects affective polarization but not the other way around.

This thesis focuses specifically on ideological polarization. In particular, the repulsive force of exposure to opposing beliefs and opinions seems to be a significant factor in understanding political polarization on social media [4]. There may be other effects present however, that can affect political opinion, and these should be considered.

## 2.6 Opinion Dynamics

The field of opinion dynamics seeks to model the spread and evolution of beliefs and opinions within a population. It combines sociological and psychological studies with mathematics to produce simulations based on both real-world data and theory. Typically, models in opinion dynamics require simplification and several assumptions to simulate a particular effect.

A well-known model within opinion dynamics is that of the Bounded Confidence Model (BCM) which describes the interaction between two agents with different opinions. Suppose we have users $a$and $b$ with opinions $o_a$ and $o_b$. The BCM posits that assuming some threshold $\epsilon$, if the distance between the two opinions is within that threshold - $|o_a - o_b| < \epsilon$ - then the opinions of the agents change according to the functions in Equation 2.9 in which $\mu$ is the strength of the shift in opinion. This mechanism suggests that upon interaction of two agents, opinions tend to move towards the center point of the pair.

$$o_a = o_a + \mu(o_b - o_a)$$
$$o_b = o_b + \mu(o_a - o_b)$$

(2.9)

The BCM has been used in political opinion focused studies along with polarization effects [7, 39].

# 3 | Simulation

Since a common approach to training RL-based RS is to use simulation, an artificial environment was devised. Several assumptions are made about this simulation with the first being that the environment represents a social media and news platform in which there are users who are recommended and interact with content. The content can be thought of as news articles or posts. The interactions between users and content could be any form of engagement such as commenting, “liking”, or reading but in the context of the simulation will simply be clicks or non-clicks.

To simulate these user-content interactions, models of both users and content were needed. The goal behind these models was to incorporate properties, which shown in research, exist in real-life political environments, to best and most realistically create an environment in which the dynamics of shifting user political beliefs can be observed. In a truly realistic scenario, the features and properties of these models would likely be based on real data. Due to the lack of a real dataset, assumptions about the user and content models were made which could be further modified given real user data.

## 3.1 User Belief

In creating a model for users, the most important consideration was to parameterize political belief in a way that it could be evolved over time with exposure to political content. Previous works and research have used a myriad of approaches to model political belief and opinion dynamics.

In Evans & Kasirzadeh, each user was given three parameters corresponding to their propensity for consuming left, center, and right wing content [14]. This approach was decided against due to wanting to model more directly the true, core political beliefs of users with a single variable. Thus, an approach like that taken in Sabin-Miller & Abrams [39] was chosen. This approach uses a single parameter $u_b$ on a continuous spectrum from -1 to 1 to model user belief where a negative score indicates a more left wing ideology, and a positive score indicates a more right wing ideology. Using a single axis for political belief is supported by studies on the political climate in America as shown in Hare and Poole [18] along with other research on political beliefs and opinion dynamics [39].

In this simulation, when a user is generated, their belief is sampled from a bimodal distribution given by Equation 3.1.

$$u_b \sim \begin{cases} \mathcal{N}(\mu_L, \sigma_L^2), & \text{if } p_L \\ \mathcal{N}(\mu_R, \sigma_R^2), & \text{otherwise} \end{cases} \tag{3.1}$$

This sampling distribution was chosen given research from Pew Research Center and Bail et al. in which the findings both show a bimodal distribution of left vs right (Democrat vs Republican) ideologies. In this distribution function, $p_L$ represents the probability a user will be sampled from the left wing distribution $N(\mu_L, \sigma_L^2)$. Values for $\mu_L$,$\sigma_L$ and $\mu_R$, $\sigma_R$ were chosen also according to the same research which found the left wing distribution to be more right skewed and the right wing distribution to be flatter. Additionally, $p_L$ should be higher than 0.5 given the ratio of Democrats vs Republicans. The simulated user population for this thesis users the values in Equation 3.2.

$$\begin{aligned} \mu_L &= -0.5 & \mu_R &= 0.3 \\ \sigma_L &= 0.25 & \sigma_R &= 0.3 \\ & & p_L &= 0.55 \end{aligned} \tag{3.2}$$

This gives the an example distribution shown in Figure 3.1 for a simulated population of 100,000 users.

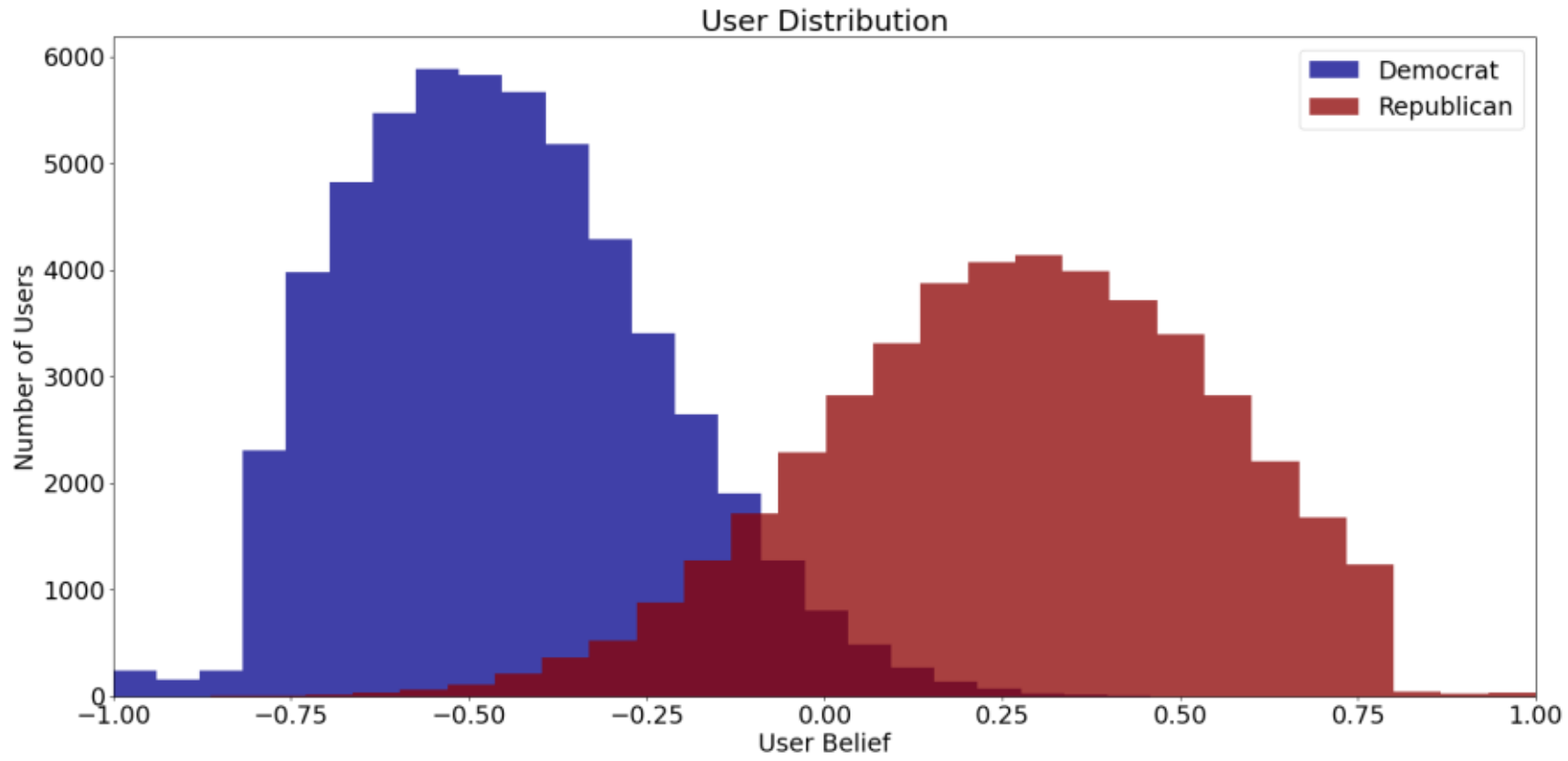

Figure 3.1: Bimodal distributions of user belief scores in population of 100,000 users

It should be noted that the Gaussian distributions were both pseudo-truncated by resampling if a sampled belief score was less than -0.8 or greater than 0.8. This was done to prevent the appearance and abundance of “extreme” users that would occur if belief scores were explicitly truncated to [-1, 1] and to assist with the asymptotic nature of belief at both ends of the spectrum. Additionally, although included in the graph, whether a user was sampled from the Democrat or Republican distribution has no further bearing within the simulation.

## 3.2 Content Bias

To model simulated content found on a social news or media platform, there needed to be both an unobservable, continuous value representing the true political nature of the content and an observable, discrete value representing the perceived political nature of the content. News articles or social media posts typically have some degree of political bias. One may be able to perceive the general extent of that degree such as knowing whether an article is written from an extreme right wing versus a more neutral viewpoint. However, each piece of political content likely falls on a continuous spectrum in terms of its bias, similar to the political belief of users.

This was incorporated into the simulation by giving each content a bias parameter and a politics parameter. The bias parameter $c_b$ exists within the same spectrum used for user belief, that is, between -1 and 1 with negative values indicating more of a left leaning bias and positive values more of a right leaning bias. The politics parameter $c_p$ is an integer in [1, 2, 3, 4, 5, 6, 7] and corresponds to a category of political viewpoint. This is slightly inspired by the work in Bail et al. which broke down the political spectrum of Twitter tweets into seven buckets. The specified categories, shown below as associated colors, are created by uniformly dividing the range [-1, 1] and respectively are named far left, left, lean left, center, lean right, right, and far right. It should be noted that although the term “center” is used, this category can also be thought of as “neutral”.

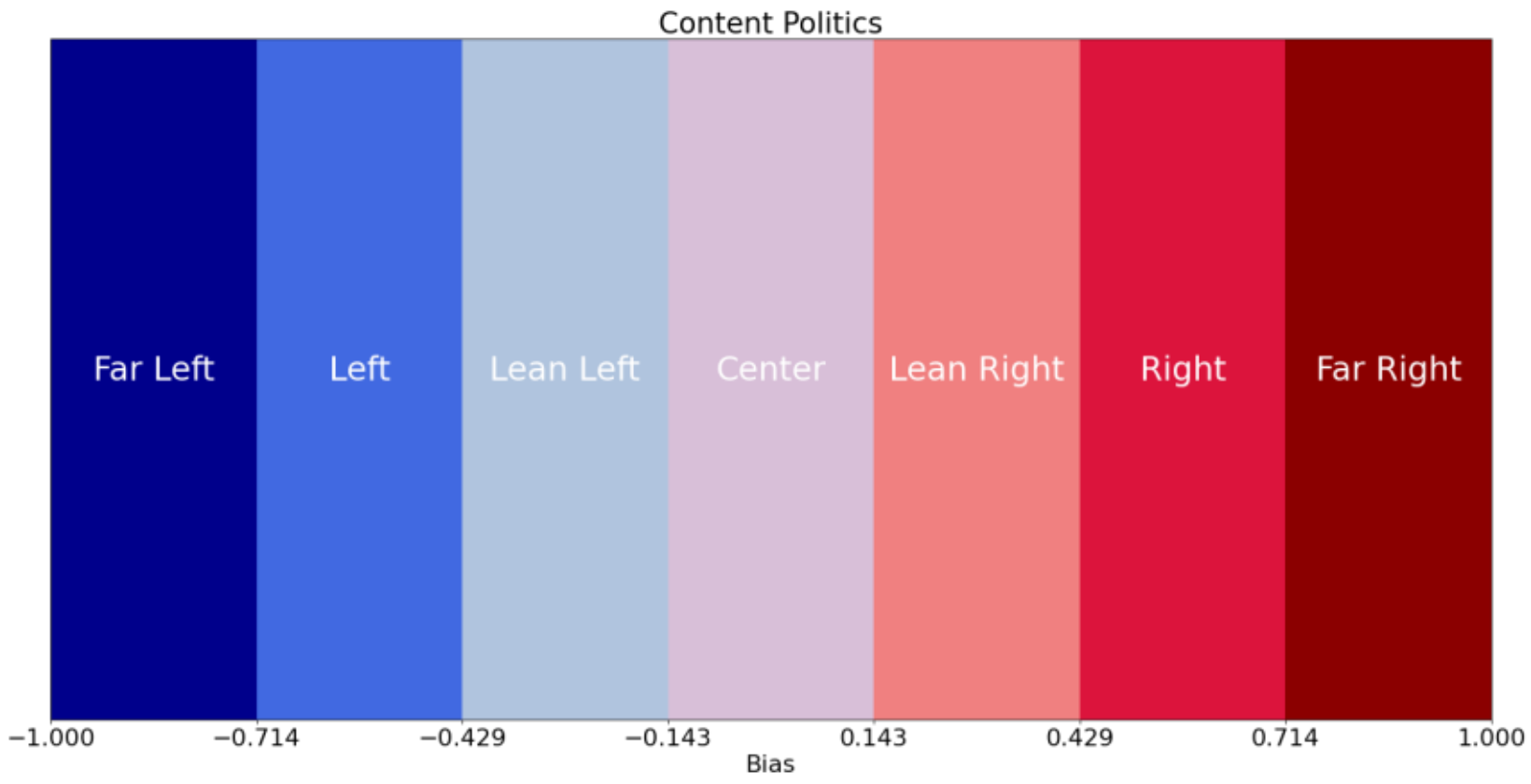

Figure 3.2: Content bias and politics category thresholds

By having a discrete value for each content, reinforcement learning agents will have a simple and restricted action space. Despite RL algorithms being capable of handling continuous action spaces where the continuous content bias parameter to be used instead for action selection, it was decidedly more realistic to use this categorical encoding approach. In reality, it would be difficult if not impossible to know the true ground state bias of a piece of content versus labeling a piece of content with one of the seven categories (i.e., via an ML classifier). Further, it’s unlikely a piece of content for every single bias value from -1 to 1 would even be available for an agent to choose from in a real-world implementation. However, it’s instead likely a social news or media platform would almost always have content from each of the seven politics categories available to be shown to a user.

When a simulated content is generated, a politics category must be provided. Using the given politics category and associate encoding, the content’s bias is sampled uniformly according to Equation 3.3.

$$c_b \sim \mathcal{U}([p_{start},\ p_{end}])$$

(3.3)

Where $p_{start}$ is the left side bias cutoff and $p_{end}$ is the right side bias cutoff for the given politics category. For example, if a far left content is requested, $c_b$ is sampled from [-1, -5/7]. This uniform sampling incorporates the inherent uncertainty there may be in categorizing a piece of content. Even though a content may seem to be neutral, it may be right leaning in terms of its true bias score, for instance. To further incorporate this uncertainty, there could be some overlap between the uniform distributions but for the sake of complexity, the simple hard cutoffs were chosen for this simulation.

## 3.3 User-Content Interactions

### 3.3.1 Opinion Shift

One of the main assumptions of this thesis is that there exist mechanisms for political opinions to shift in both directions on the single axis political spectrum. In Evans & Kasirzadeh [14], the effect reported in Bail et al. was implemented which causes exposure to opposing views to shift a user's opinions in the opposite direction. This polarization effect seems to be supported by other research as shown in [11, 26]. To complement this repulsion effect, a mechanism of attraction was also incorporated which is supported by the Bounded Confidence Model (BCM).

Several parameters were devised to model the opinion shift dynamic. The first of which is called the *dissonance* factor $d$ and is given by the difference between content bias and user belief, or $d = c_b - u_b$. This allows for the shift in opinion to change according to how close or far away the bias of a piece of content is from the user's belief.

Another parameter, the *polarization factor* $u_{pf}$ is a user feature which describes the degree of which polarizing content affects their opinion. This along with the *dissonance* factor are both borrowed from the work in Sabin-Miller & Abrams in which the two factors are the main components of opinion shift. By combining the two factors in a similar manner the function in Equation 3.4 can be derived:

$$u_b = u_b + \frac{d(1-d^2)}{u_{pf}^2} \tag{3.4}$$

This allows for user belief to shift in the same direction of content bias when the *dissonance* is small but move in the opposite direction instead when the *dissonance* is large.

The opinion shift function was further expanded with the addition of two multiplicative factors - *malleability* and *engagement*, and *extremes decay* - which all may increase or decrease the opinion shift. The first, *malleability* is a user feature $u_m$ which is defined as the influenceability or how malleable a given user's opinions are. This value may be the same for all users or distributed in some way across a spectrum.

Another user feature, *engagement* $u_e$ was created to help make the simulation more realistic. Each user starts with $u_e = 0$ and when a user engages (i.e., clicks) with a piece of content, the user's *engagement* will increase by some small value up to an upper bound of 1. What this hopefully captures is the difference in opinion shift strength for a user who is actively engaged on a platform vs a new user who is being served content for the first time and may not have "trust" yet for the platform.

Lastly, the *extremes decay* $e_d$ factor seeks to enforce the asymptotic nature of the extremes on the belief spectrum. This factor is given by the piecewise function in Equation 3.5.

$$e_d \sim \begin{cases} 1 - u_b^2, & \text{if } d(1-d^2) \cdot u_b > 0 \\ 1, & \text{otherwise} \end{cases} \tag{3.5}$$

The sign of $d(1-d^2)$ determines the direction of the opinion shift - to the left or right. Thus, if the product of this value and user belief are positive, the shift is pushing the user's opinion towards the respective extreme on their side of the political spectrum (i.e. a left wing user's opinion is shifted more towards the left). In this case, the shift is reduced proportionate to user belief by $1 - u_b^2$ so that the closer to the extremes on the left or right a user's opinions are, the more difficult it will be to continue to push them in that direction.

This overall helps to mitigate user opinion from reaching and piling up at the extremes of -1 and 1.

Bringing all these pieces together, Equation 3.6 shows how user belief evolves in the simulated environment.

$$u_b = u_b + \frac{d(1-d^2)}{u_{pf}^2} \cdot u_m \cdot u_e \cdot e_d \tag{3.6}$$

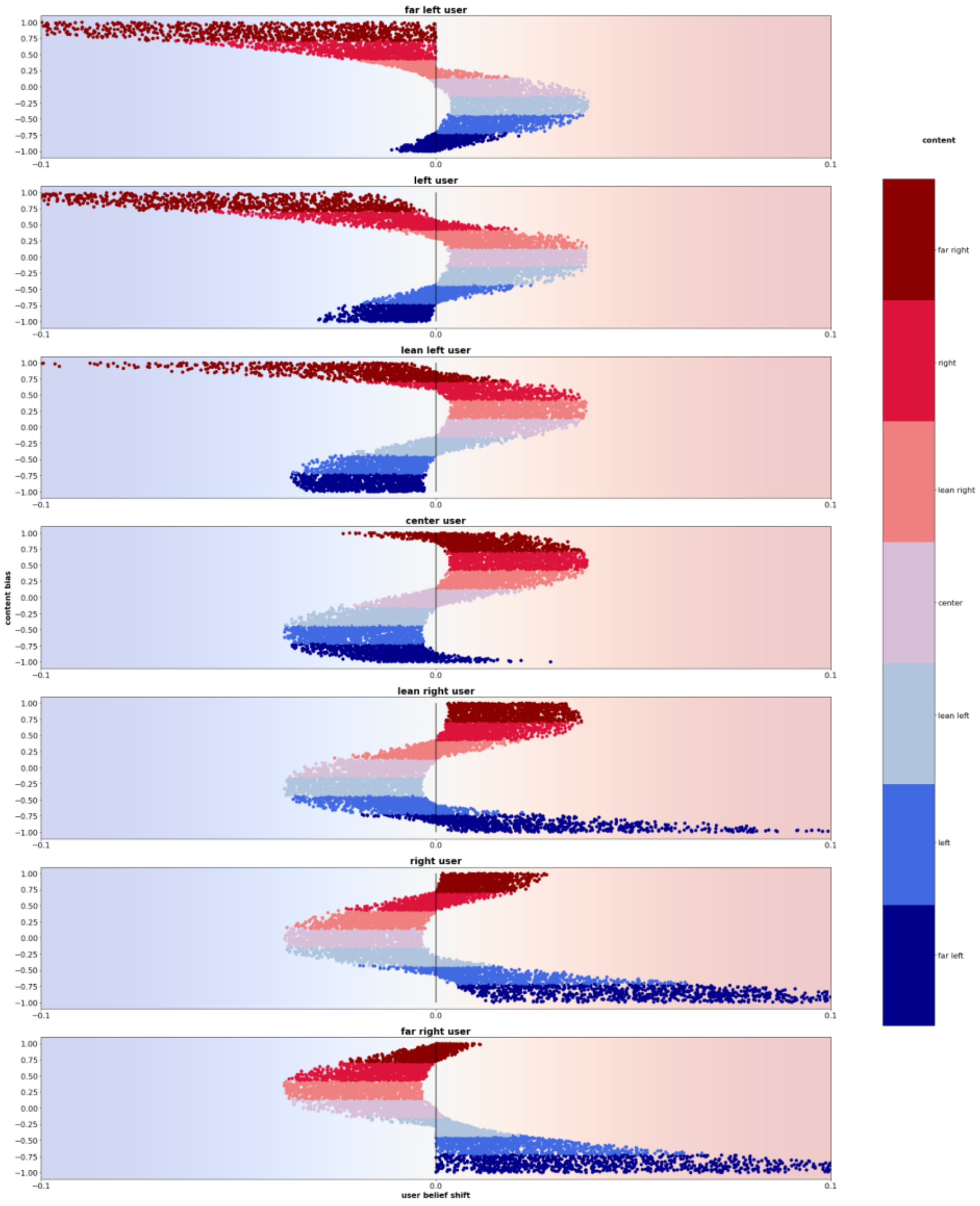

Figure 3.3: Average belief shift per user politics type

This equation can be visualized by sampling users and comparing the shifts in opinion when recommended content from each political category as shown in Figure 3.3. Here we can see that in general, being shown content with opposing bias will embolden a user's belief more towards the extremes. Conversely, being shown content with bias closer to a user's belief will instead shift the user's belief in the direction of the bias. For example, a lean left user shown center content will have their belief shifted slightly to the right.

### 3.3.2 Click Probability

Since user-content interactions consist of whether a user clicks on a recommended content, a way of producing click probabilities was required. Here, several liberties were taken due to the lack of real data, about how the probability distribution might look for a given user and content. In general, it was assumed that users are more likely to click content with bias closer to their own belief scores. It was also assumed that click probability falls off as the dissonance between belief and content bias grows, but that a user still has some non-zero probability of clicking content with bias slightly to the left or right of a user's belief.

To produce probability distribution fulfilling these assumptions, a few additional parameters were created. The first is *open-mindedness* which is a user feature $u_{om}$ and can be described as how open-minded a user is to content with bias farther from their own belief. In terms of click probability, it affects how horizontally stretched the click probability distribution is.

Next, two parameters were added to the interaction function to allow for fine-tuning of the click probability distributions. The first is the *probability spread* $p_{spread}$ and as the name suggests, controls the spread of the probability distribution. Higher values cause the "bell" of the distribution to be wider. The second parameter is *max probability* $p_{max}$ and simply controls the highest possible click probability.

Now, the click probability for any given user and content is sampled according to Equation 3.7.

$$p_{click} = \sigma\left(\frac{u_{om}}{|d| \; + \; 1e^{-8}}\right)^{p_{spread}} \cdot p_{\max} \tag{3.7}$$

Here, $\sigma$represents the sigmoid function which restricts the probabilities to [0, 1]. The dissonance factor from the opinion shift function is also included but the absolute value is taken. A small non-zero value, $1e^{-8}$ is added to the denominator within the sigmoid function to prevent division by zero in the case that dissonance is zero. Admittedly complex, the click probability function does produce the desired probability distributions as shown in Figure 3.4 and can be modified in a variety of ways by tuning the parameters.

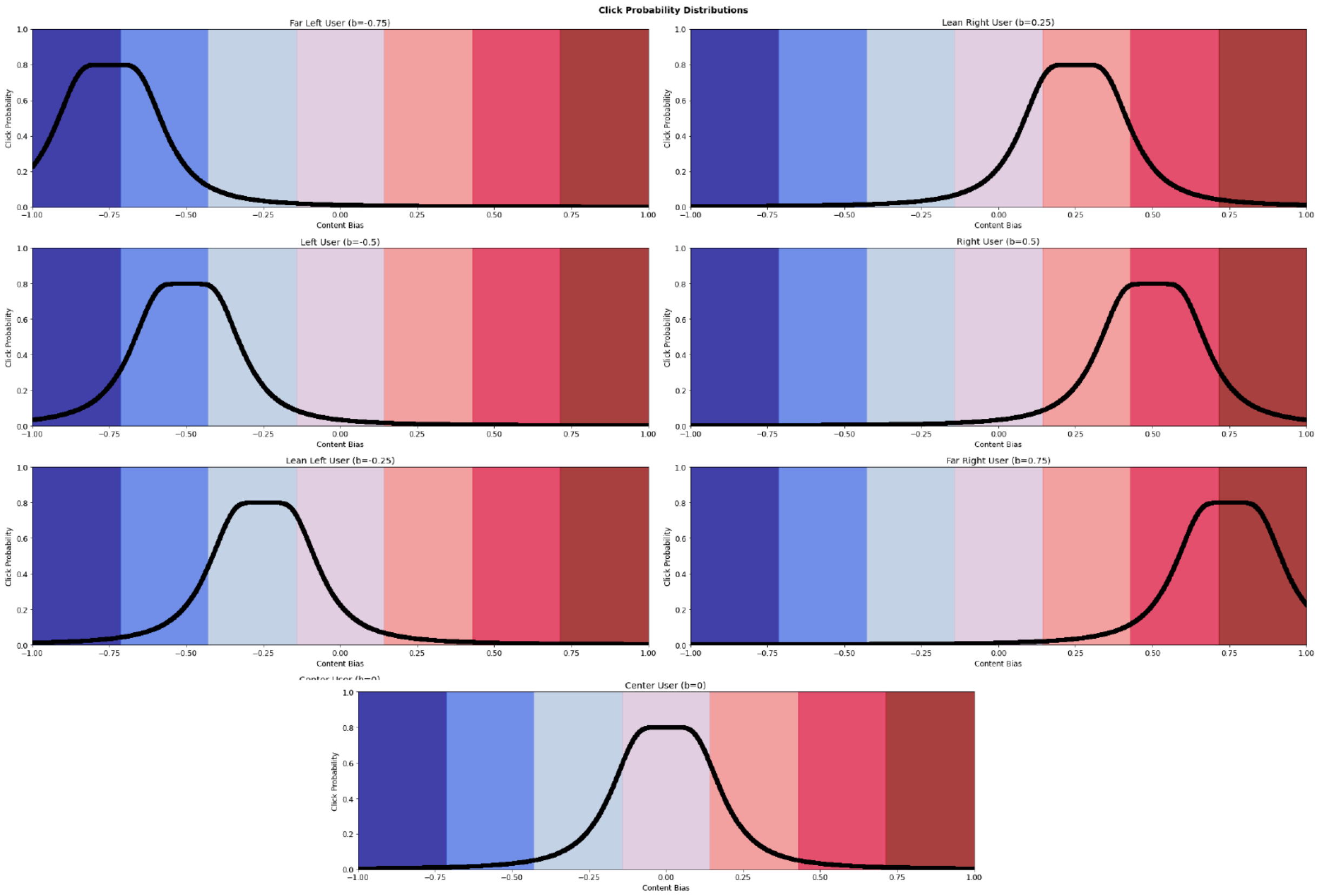

Figure 3.4: Click probability distributions across user belief and content bias spectrums

### 3.3.3 User Attrition

In real world environments, if a user is served poor recommendations for too long, their satisfaction with the platform is likely to decline and they may eventually choose to leave the platform entirely [38]. To incorporate this dynamic in the simulation, a user attrition mechanism was added. Attrition is also known as customer churn typically in a sales context in which customers are lost over time but can occur in any platform.

For this mechanism a parameter called *satisfaction* was added to the user model. This parameter is bound to [0,1] and is initialized to 1 when a user is created. When a user interacts with a piece of content, the *satisfaction* parameter is decayed if the user did not click on the content but increased if the user did click on the content. These growth and decay rates are sampled uniformly between 1.01 and 1.10. If a user's *satisfaction* falls below a chosen threshold, there is a chance for attrition. This attrition chance is given by the piecewise function in Equation 3.8.

$$p_{attrition} \sim \begin{cases} 1 - \frac{u_{satisfaction}}{threshold_{satisfaction}}, & \text{if } u_{satisfaction} < threshold_{satisfaction} \\ 0, & \text{otherwise} \end{cases} \tag{3.8}$$

To illustrate the attrition mechanism, a sampling of 1000 users were recommended content at random until all users had left the platform. This shows in Figure 3.5. that on average, users may start churning after ~30 random recommendations and by ~80 random recommendations, all users have churned.

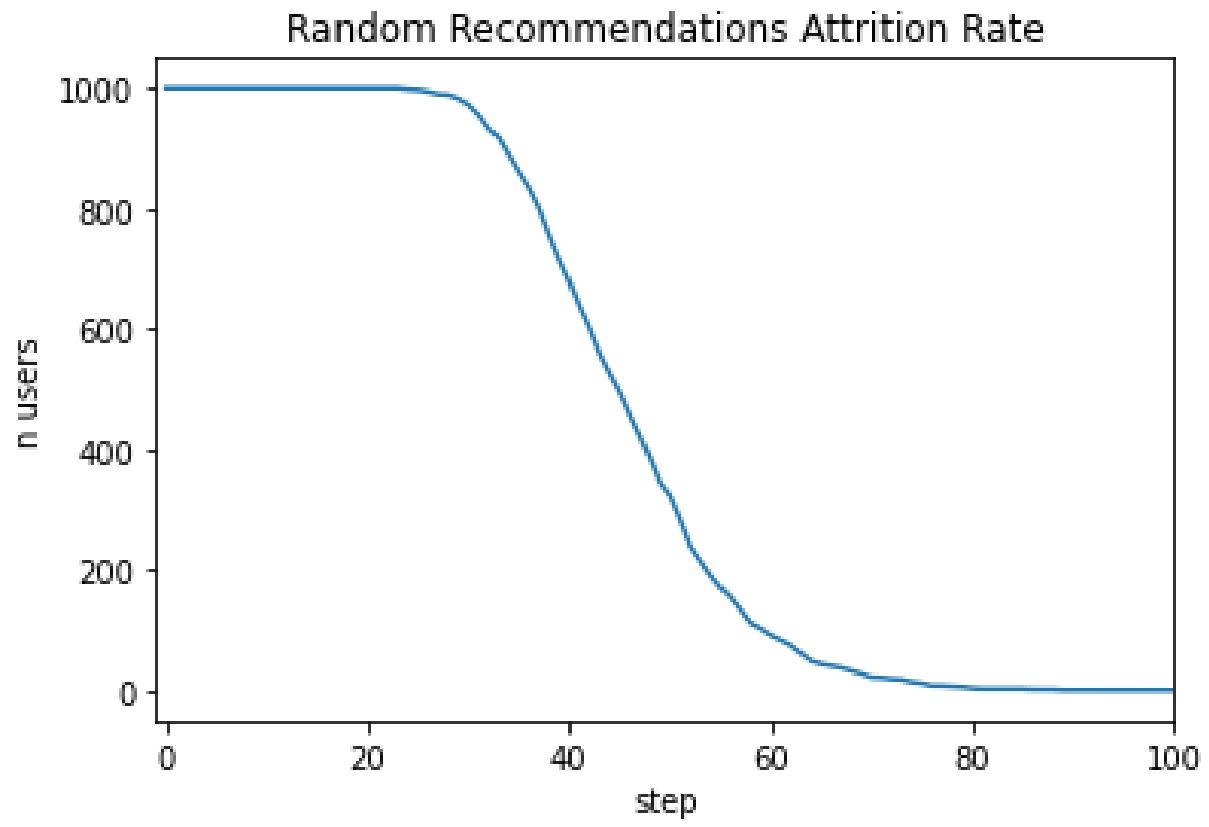

Figure 3.5: Attrition rate for 1000 users served randomized recommendations

# 4 | Experimental Setup

## 4.1 Environment

The requirements of the environment included representing information on the types of content shown historically to users along with whether the user clicked on them. Additionally, a simple and realistic reward function was needed along with a mechanism of episode termination.

### 4.1.1 Observations

Gym is an open source Python library which provides tools for interfacing environments with reinforcement learning agents. A custom Gym environment was created which incorporates the simulated users, content, and interactions. This environment begins an episode by sampling a user from the bimodal distribution described in the previous section. The observations provided by the environment consist of a 2D array that contains information on the previous actions taken. The observations are of dimension *horizon* x 8 where *horizon* is the history length (i.e., how many previous interactions are considered). Each row consists of a one-hot encoded representation of the content politics (the first seven values) and whether the content was clicked. For example, a user who has been served a piece of left content and clicked on it would have the row shown in Figure 4.1 in their history.

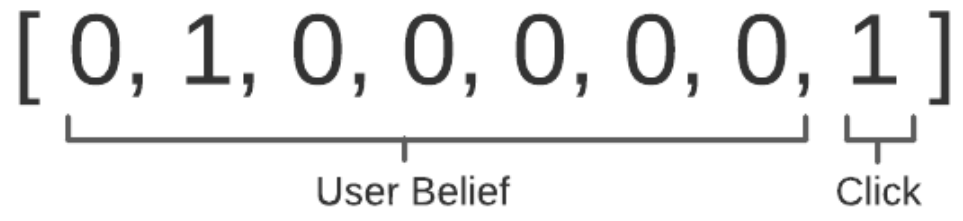

Figure 4.1: Example observation row

By choosing to represent a user's click history in this way in the agent's observations, the agent can develop a sequential understanding of its actions and how they affect the latent states of the users. A *horizon* of 100 was chosen for the experiments as that allows enough time and recommendations shown for a notable change in user belief.

### 4.1.2 Rewards

The reward function was kept simple in that a click rewards the agent +1 points. If the user attrits, a reward of -1 is given but no negative reward is given if the user reaches the end of the episode. In the experiments, this reward function is changed in the case of encouraging manipulation.

### 4.1.3 Termination

The environment reaches a terminal state once either the user is no longer active (i.e., they attrited) or a predefined *user lifespan* was reached. Although in a real world RS environment, the only terminal state is when the user leaves the platform, a somewhat arbitrary number of timesteps was chosen so that episodes could be finite.

## 4.2 Model

A DQN architecture was chosen due to its use and success in other RL-based RS and the reinforcement learning library RLlib was used to build and train the model. RLlib provides production ready implementations of many RL algorithms with an emphasis on scalability and parallelism. The latter benefit was especially important in running the experiments of this thesis due to time and compute limitations. Specifically, the Apex-DQN algorithm was used due to its ability to train quickly in a distributed fashion. Apex style algorithms utilize multiple actors or agents who interact with different instances of the environment, choosing actions according to one central policy neural network. After experience has been collected, only the most important experiences are focused on during training. This allows for a high volume of experience collection and exploration, without wasting time training on “easy” or unhelpful data.

## 4.3 Training

Training for all experiments was done using a Google Colab notebook. Google Colab provides cloud based Jupyter notebooks powered by a GPU. Specifically, using a Colab Pro+ account, training was done on a notebook with 8 CPUs and 1 NVIDIA Tesla T4 GPU. Each CPU was designated as a worker in RLlib with 100 environments each. By giving each worker many environments, action predictions by the model could be done in batch, helping to speed up experience collection and training.

## 4.4 Experiments

For comparison of models with intentionally manipulative reward functions, three different non-manipulative models were trained.

The first is a random model which chooses recommendations at random. The second is a baseline model which has a manually created algorithm to handle recommendations. The baseline algorithm works by initially using weighted sampling of the actions given the relative observed click-through-rates of each action. After 25 recommendations, the content politics type with the highest CTR thus far is chosen at each step. This allows for some initial exploration and profiling of user preferences before solely recommending the user's most preferred content. Lastly, an Apex-DQN model is trained with a "normal" reward function that gives +1 for every clicked recommendation, +0 for non-clicks, and -1 if the user goes inactive.

Two different models are trained with manipulative reward functions. The first seeks to explicitly polarize users by rewarding only far left or far right content. The second seeks to depolarize users by reward only lean left, neutral, and lean right content. These two models help to show cases of "good" vs. "bad" manipulation.

A summary of each experiment and the difference in content click rewards can be found in Table 4.1. To evaluate each experiment, a fixed population of 1000 users was generated. The model of each experiment was tested on these users and the resulting changes in belief distributions were observed. Additionally, the average recommendations for each category of user (far left, left, lean left, center, lean right, right, far right) were calculated at each time step to help visualize the policies. Other metrics such as CTR and attrition rate were also calculated.

| Experiment | Model | Content Click Rewards | | | | | | |
|---|---|---|---|---|---|---|---|---|
| | | Far Left | Left | Lean Left | Center | Lean Right | Right | Far Right |
| Random | Random | +1 | +1 | +1 | +1 | +1 | +1 | +1 |
| Baseline | Algorithm | +1 | +1 | +1 | +1 | +1 | +1 | +1 |
| No Manipulation | Apex-DQN | +1 | +1 | +1 | +1 | +1 | +1 | +1 |
| Polarizing | Apex-DQN | +1 | 0 | 0 | 0 | 0 | 0 | +1 |
| Depolarizing | Apex-DQN | 0 | 0 | +1 | +1 | +1 | 0 | 0 |

Table 4.1: Summary of experiments and content click rewards

# 5 | Results

## 5.1 Random Agent

The random model performed as expected on the user test population. Although a small overall shift in belief distribution, the beliefs of users were relatively unchanged as shown in Figure 5.1.

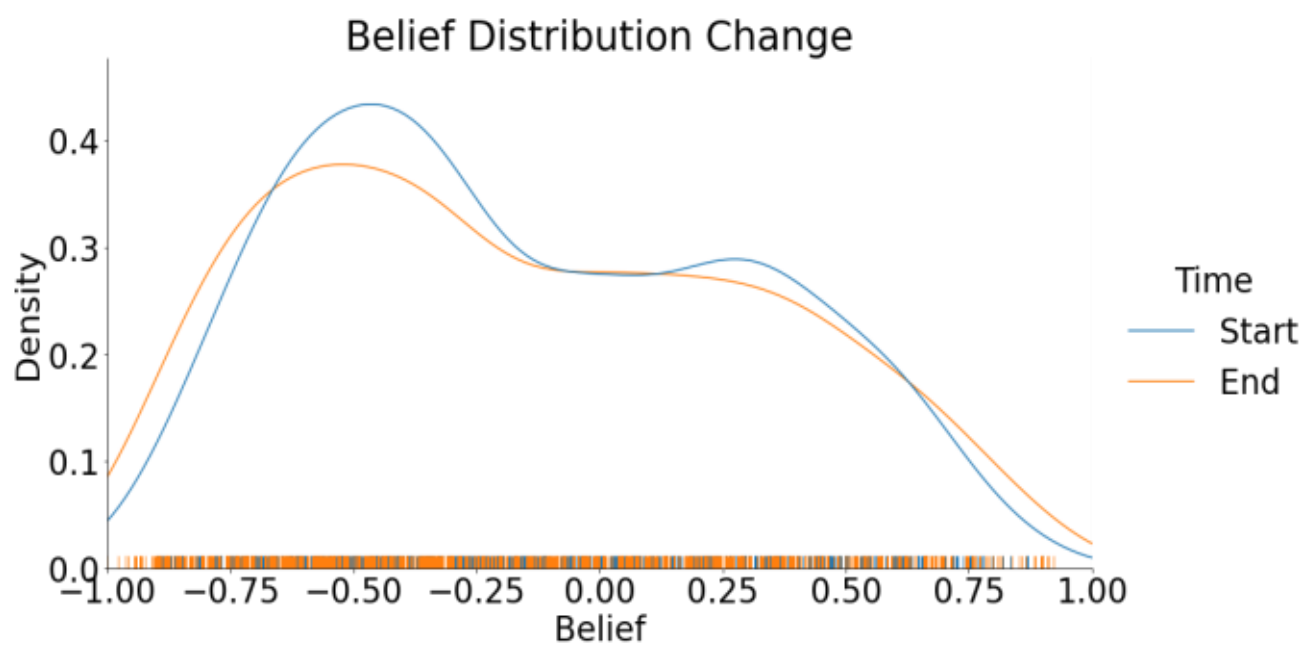

Figure 5.1: Random agent: belief distribution change

As seen in Figure 5.2, the random model was unable to stop users from churning and thus all users attrited by step 80 with the majority leaving by steps 50-60. This along with the *engagement* mechanism, prevented significant belief change which seems realistic given the assumption that users shown random content on a new platform would not have their opinions swayed until they “trusted” the platform.

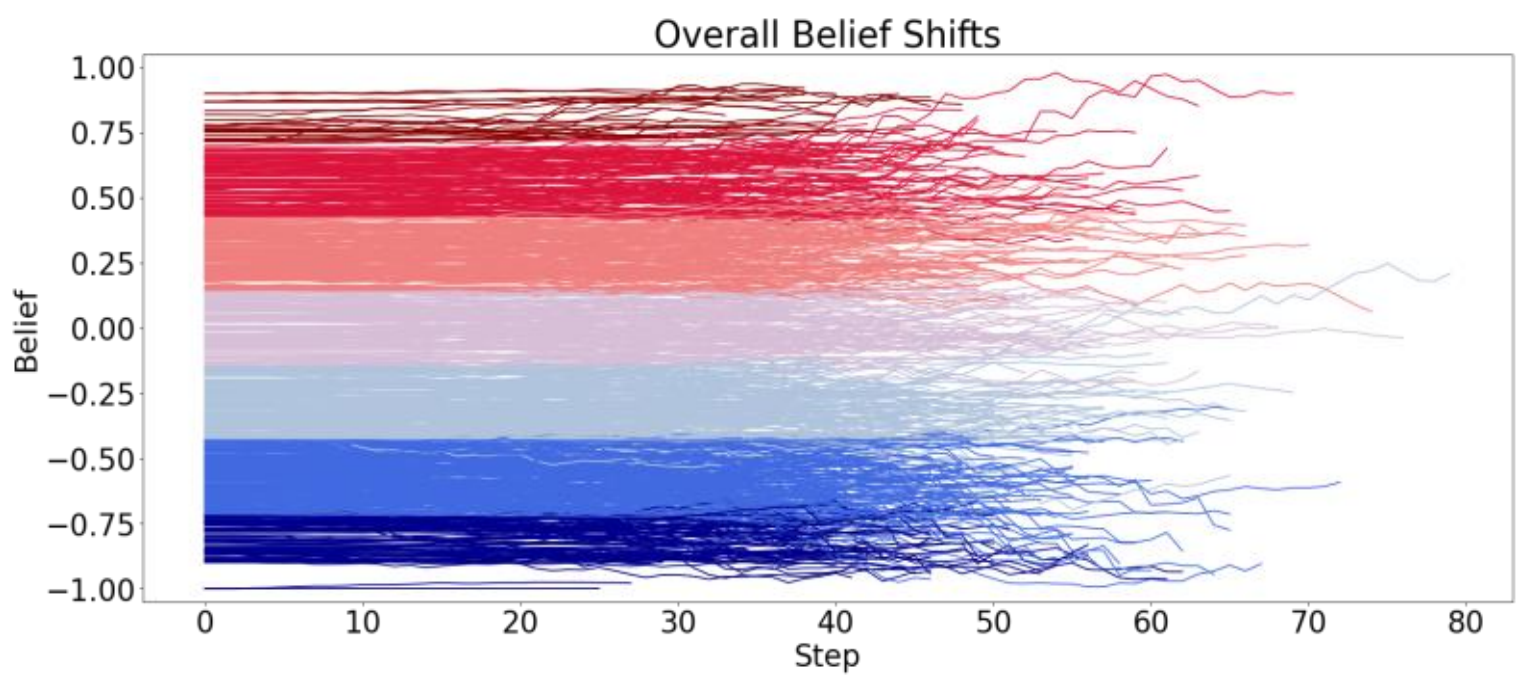

Figure 5.2: Random agent: overall belief shifts

Lastly, for comparison’s sake, Figure 5.3 shows the average recommendations for each user ideology type which of course shows a random pattern of content recommendations. The grey regions in these graphs indicate that no user survived that many steps in the environment.

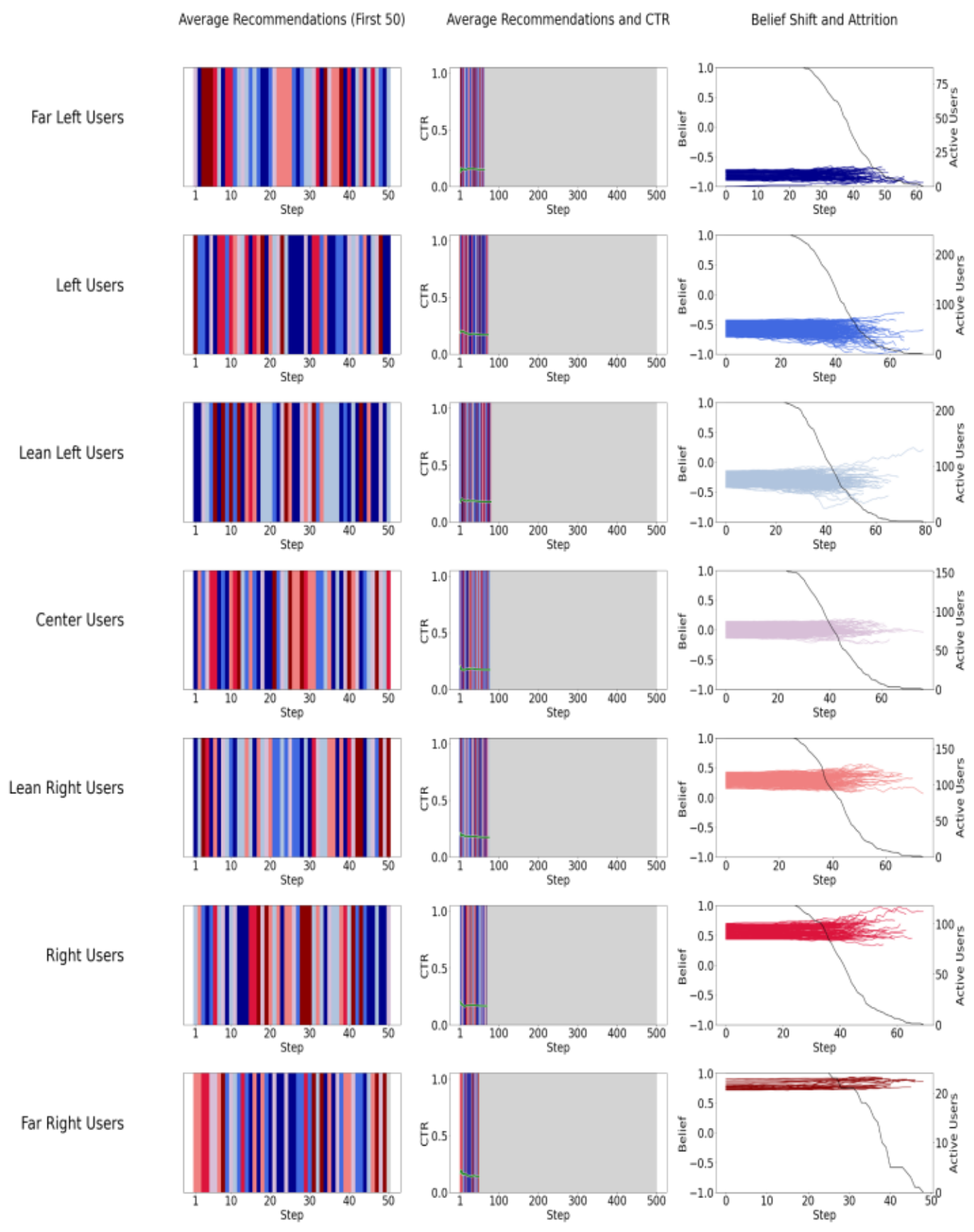

Figure 5.3: Random agent: average recommendations by user ideology type

## 5.2 Baseline Agent

The baseline model utilized a simple method of first profiling users by recommending content pseudo-randomly (with weighted probability) and then recommending only the content type with the highest CTR. This resulted in some change in belief distribution for the test population as shown in Figure 5.4.

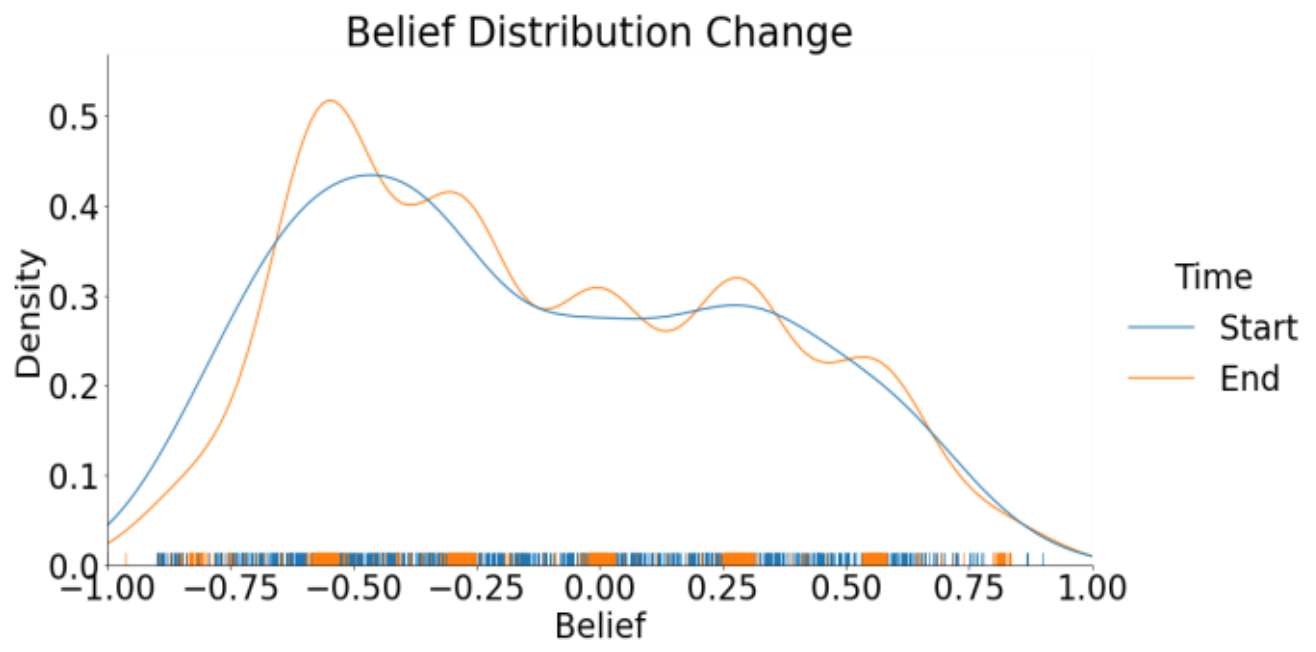

Figure 5.4: Baseline agent: belief distribution change

The baseline model was able to prevent users from leaving the platform as shown by the survival of nearly all agents to step 500 in Figure 5.5. Here, we can also see that the baseline model creates what look like echo chambers as users were condensed to seven small belief ranges.

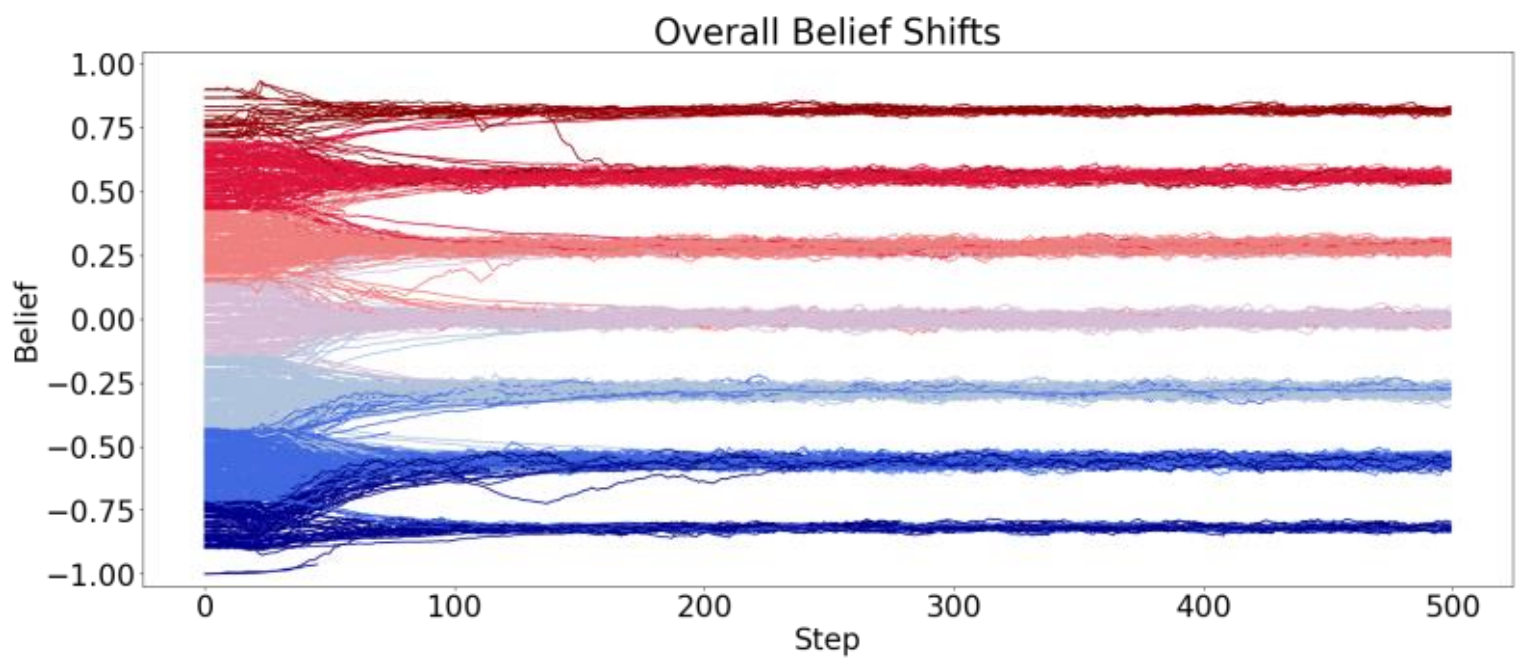

Figure 5.5: Baseline agent: overall belief shifts

In Figure 4.6 we can see the profiling strategy of the baseline model in action. For each ideology type, the baseline model was able to determine and recommend content with the highest click probability. The only exception is that of far left users in which the baseline model recommends predominantly left content instead of the expected far left content. Interestingly this does not happen with far right users and content. Still, recommending left content to far left users is somewhat successful with most users surviving the 500 steps.

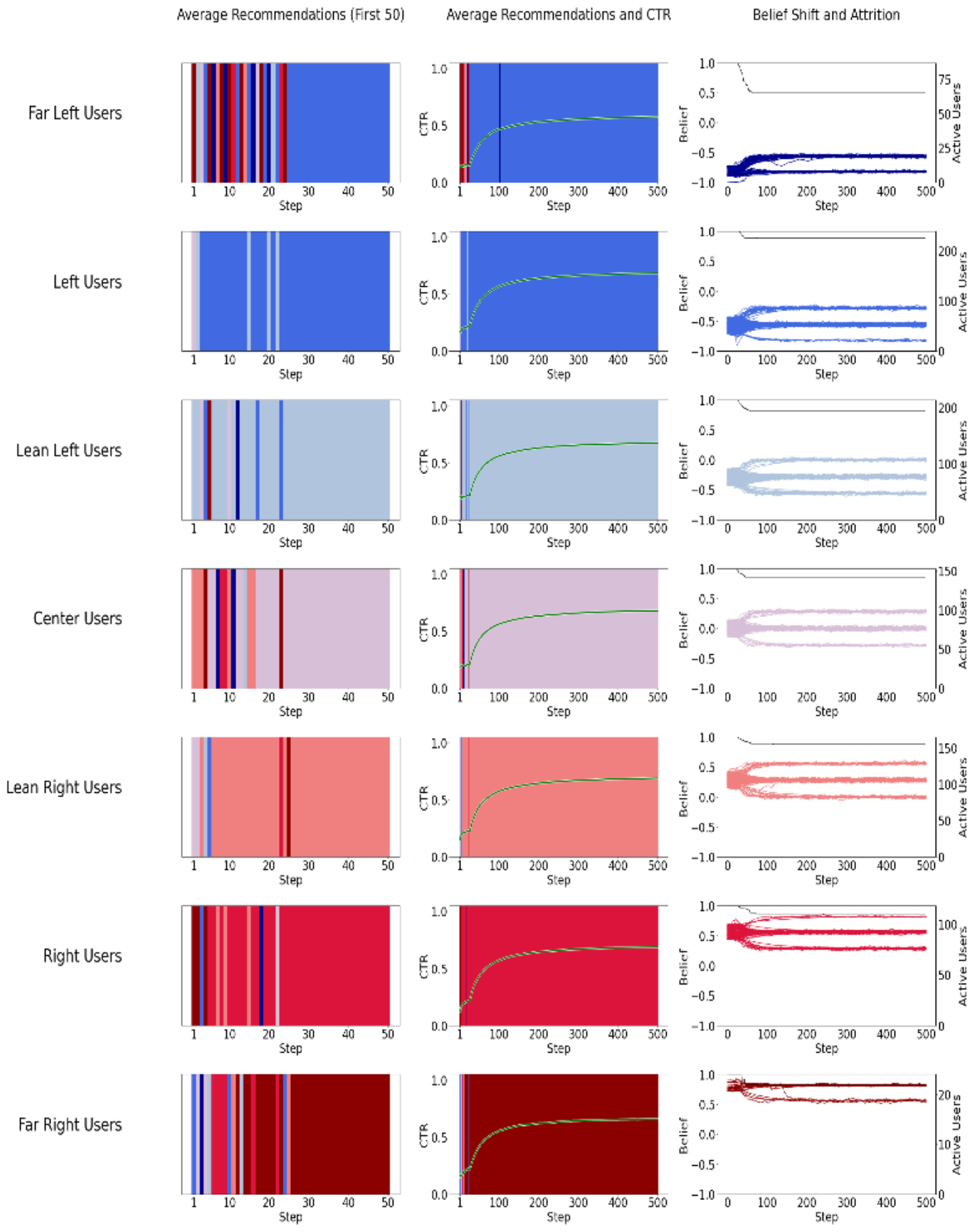

Figure 5.6: Baseline agent: average recommendations by user ideology type

## 5.3 No Manipulation Agent

The no manipulation model is the first trained model and thus training metrics can be analyzed as shown in Figure 5.7. This model was trained for 150,000 iterations after which no further improvements were noticed. Although higher initially in the training process, the average user belief shift was low for this model. Since this model has no incentive to shift user preferences, this is as expected.

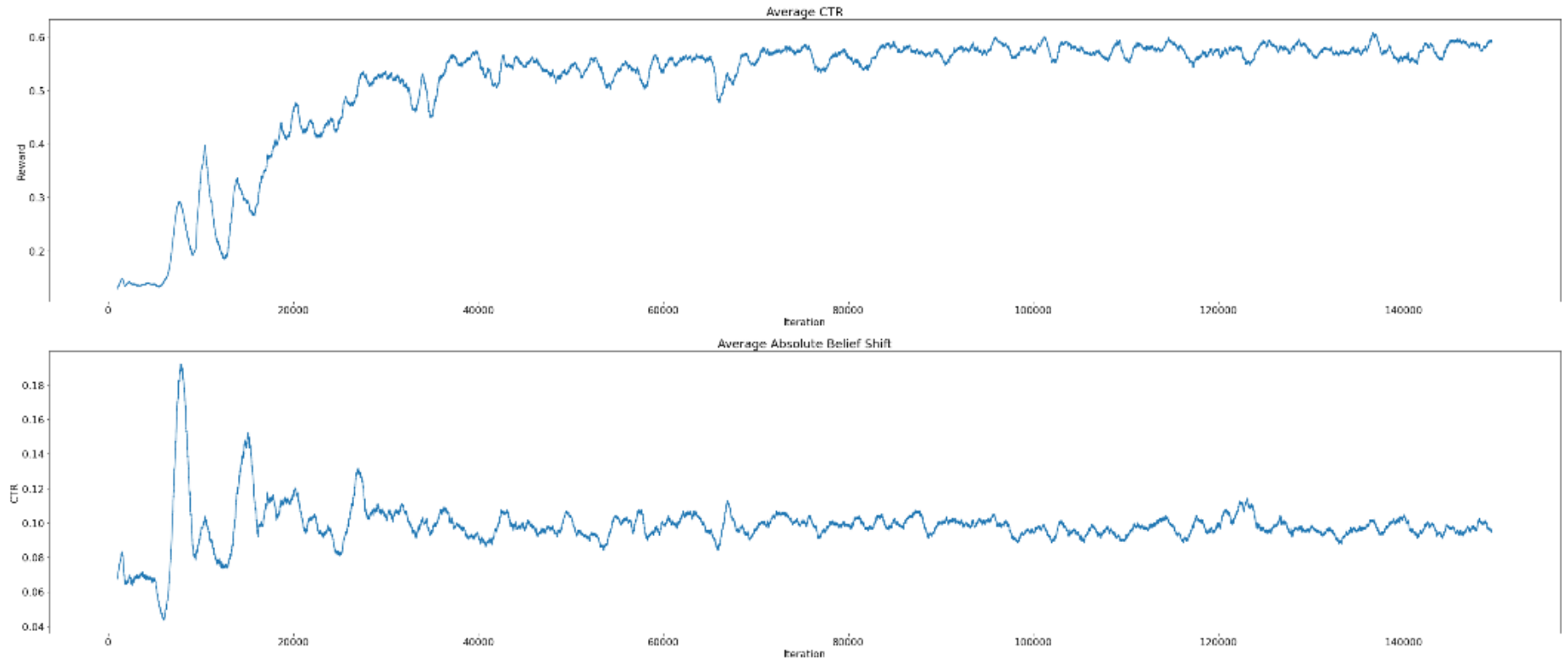

Figure 5.11: No manipulation agent: training metrics

Figure 5.12 shows the overall belief distribute change. From this, it appears that this agent has developed a policy which shifts the belief users on the left and users on the right toward two peaks, namely at -0.5 and 0.25 respectively. The peak on the left side of the spectrum however actually appears slightly bimodal. These peaks do line up with the natural peaks of the underlying user distribution, so it appears that the no manipulation agent is simply reinforcing the given distribution. This could be said to be increasing polarization, as the average distance between the belief of users is higher but we do see less extreme views overall.

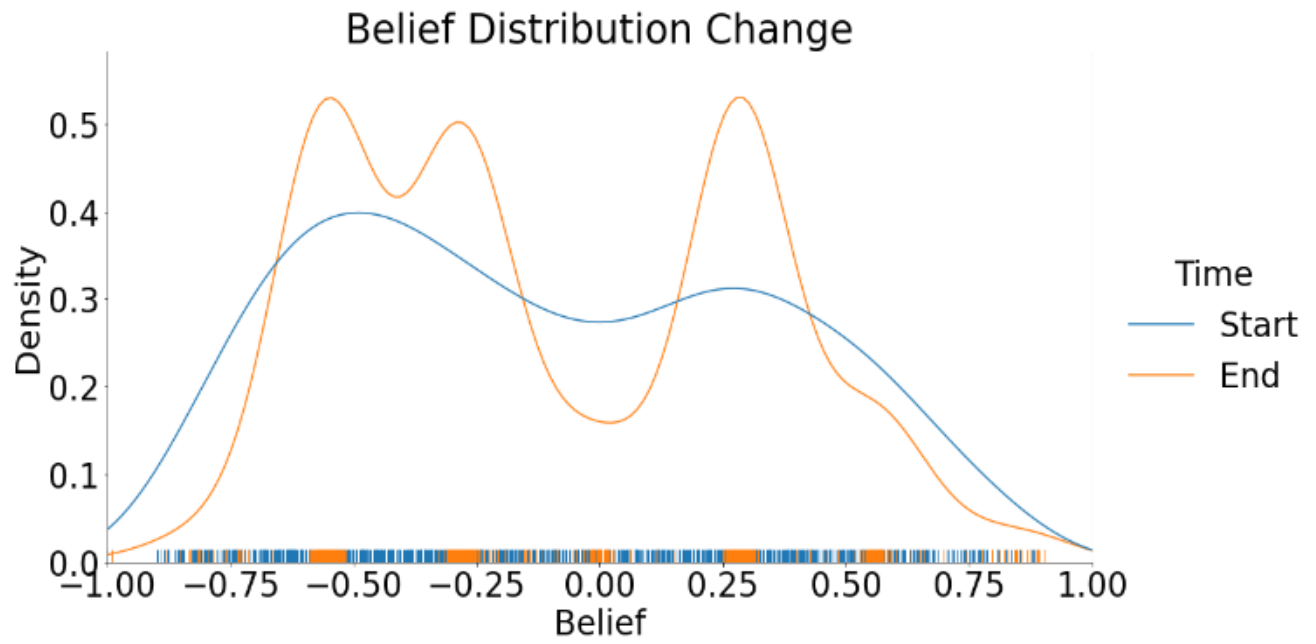

Figure 5.12: No manipulation agent: belief distribution change

In Figure 5.13, we can see that user belief was condensed similarly to the baseline agent. However, in this case, there are no far right users at the end of the recommendations. There are far left users but far less compared to the other ideological categories.

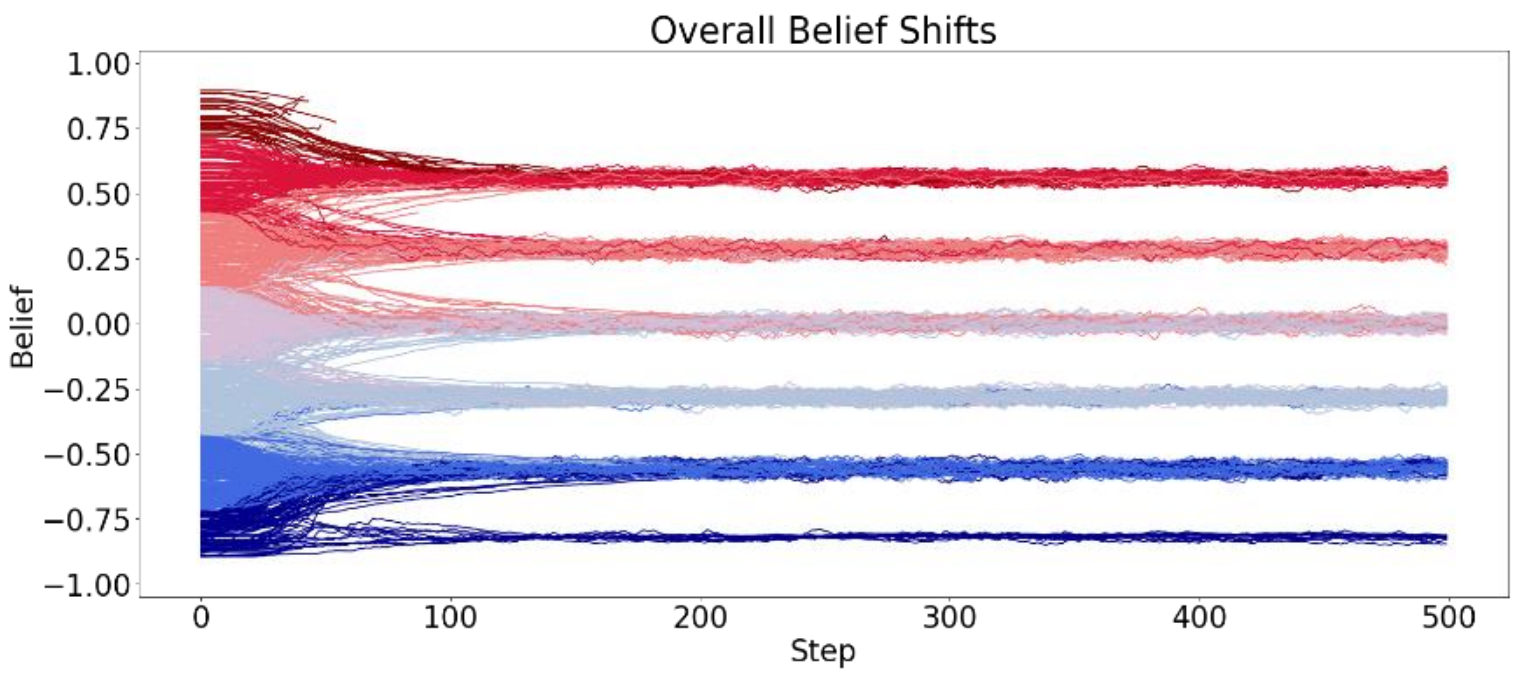

Figure 5.13: No manipulation agent: overall belief shifts

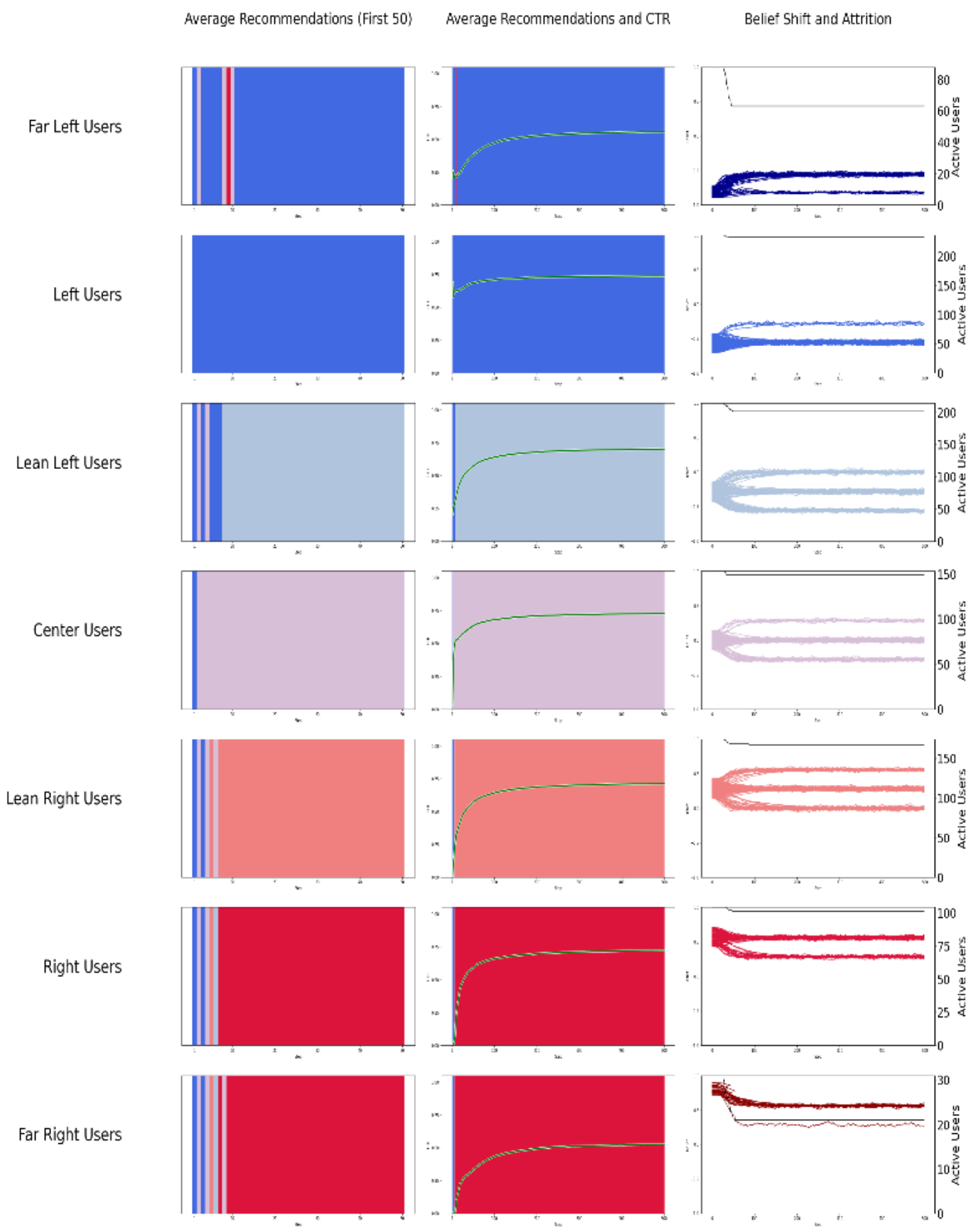

Figure 5.14: No manipulation agent: average recommendations by user ideology type

Figure 5.14 shows how this strategy played out. We can see a similar profiling strategy that was used in the baseline agent that was learned by the model. The agent here on average does not recommend any extreme content (far left or far right). Instead, it treats extreme users the same as users in the left and right categories. This seems to have caused higher attrition rates and lower CTRs for far left and far right users.

## 5.4 Polarization Agent

The polarization model required a longer training period than the no manipulation model. This makes sense given the added difficulty of needing to shift user preferences to maximize rewards. The polarization model was trained for 1,000,000 episodes and the improvement in CTR as well as average absolute belief shift can be seen in Figure 5.11. Both metrics follow a similar pattern because the two metrics are indirectly tied together via the modified reward function.

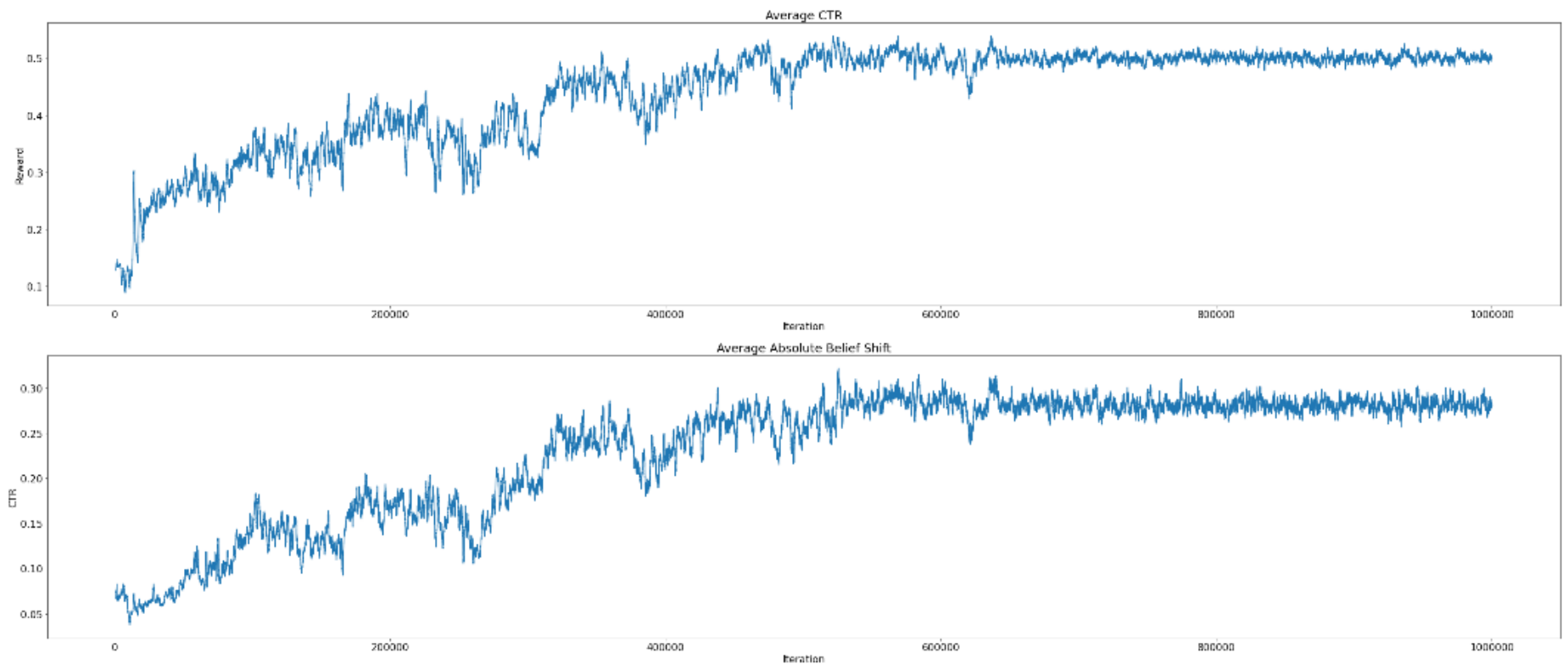

Figure 5.11: Polarization agent: training metrics

In Figure 5.12 we can see the dramatic effect the polarization model had on the user belief distribution. As intended, the model has shifted user belief to the extreme ends of the belief spectrum.

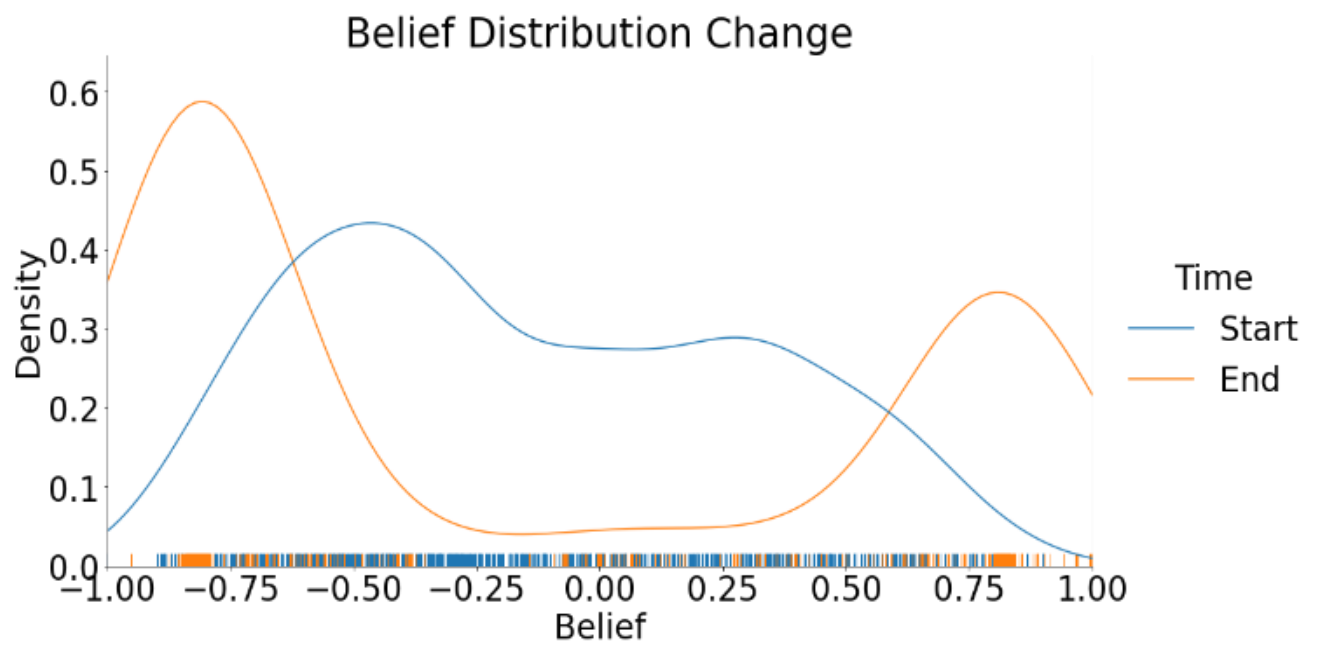

Figure 5.12: Polarization agent: belief distribution change

The overall belief shifts shown in Figure 5.13 also shows this drastic shift. Like the baseline and no manipulation models, users are condensed into small belief ranges, but in this case, only two groupings at both extremes.

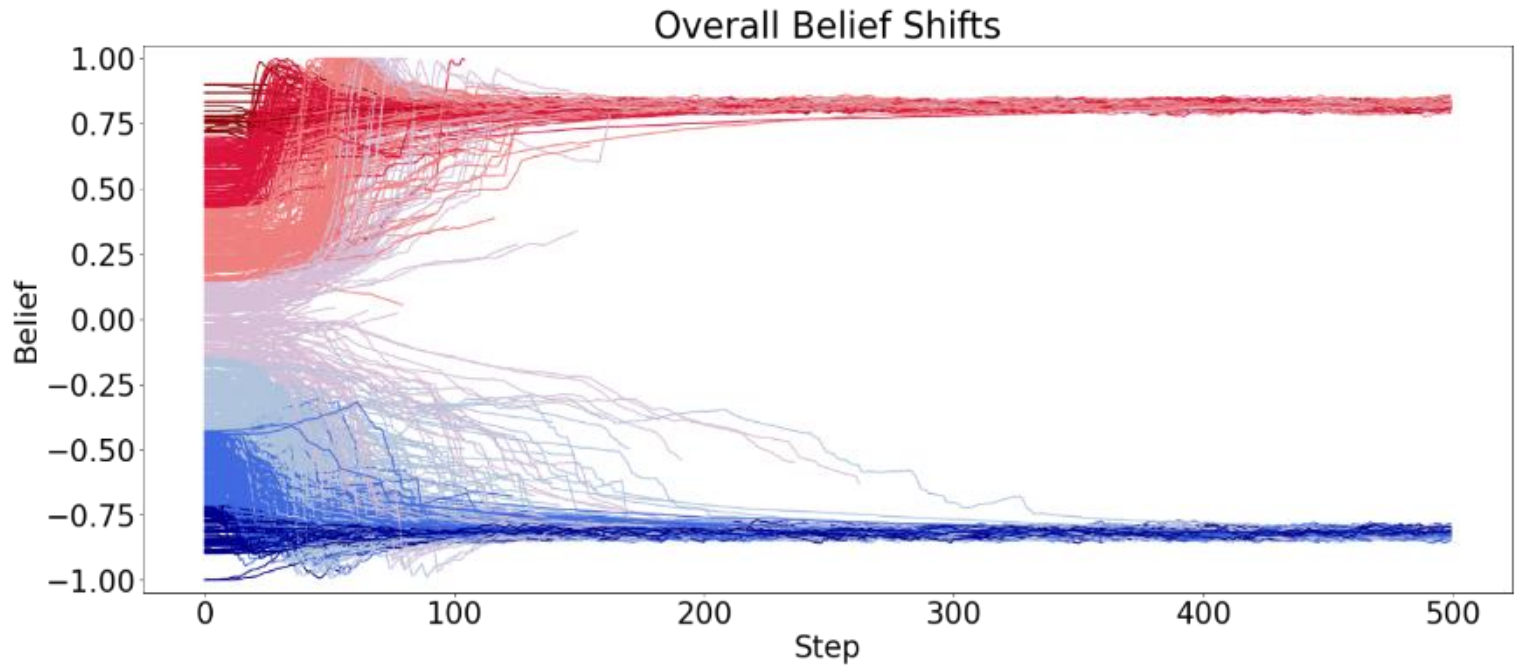

Figure 5.13: Polarization agent: overall belief shifts

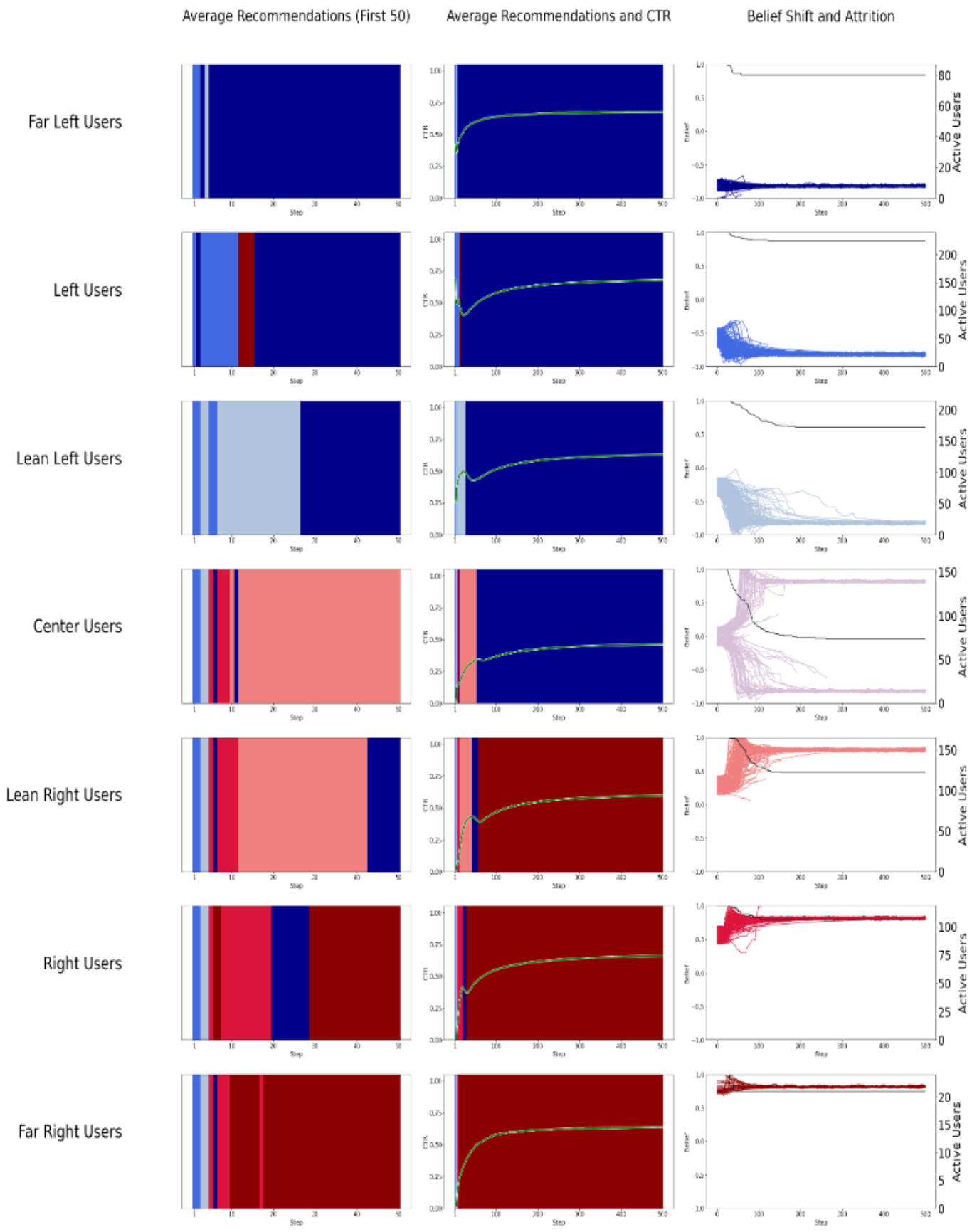

Figure 5.14: Baseline agent: average recommendations by user ideology type

In Figure 5.14, the strategy of the model to polarize users can be seen. For users in the lean left or lean right categories, the model recommends lean left or lean right content respectively for several steps and then begins recommending more extreme content. This can be seen as the model learning to gain the trust of users so that their *engagement* score rises. Then, once a user is fully engaged, recommending extreme content shifts the user's belief toward that extreme. In the lean right case, we can even see the model utilizing the repulsive polarization effect as it recommends far left content to lean right users to help nudge their beliefs more to the left.

The polarization model struggles most with center users. This is not unexpected as the belief shift required for these users is the highest. We can see that for center users, the polarization model causes an attrition rate close to 50%. The model also achieves a lower peak CTR for center users. Still, the polarization model can shift at least half of center users to one of the extremes.

One interesting element of the polarization model policy is that, even though user beliefs get condensed to a belief score of either ~-0.8 or ~0.8, often user belief overshoots these points initially before shifting back towards them. This is likely due to differences in the content served leading up to the point at which the model recommends only far left or far right content. Because of the random nature of clicks, the model may be essentially "playing it safe" by recommend more intermediary content to ensure that the user's belief has reached the point of being interested in more extreme content.

## 5.5 Depolarization Agent

The depolarization model was also trained for 1,000,000 episodes as shown in Figure 4.15. Similarly, to the polarization model, CTR and absolute belief shift are indirectly related but in the case of the depolarization training metrics, we do not see coupling of the two the same extent. This may be due to a larger proportion of users initially starting with belief scores that make them receptive to more neutral content and thus less overall belief shift is required.

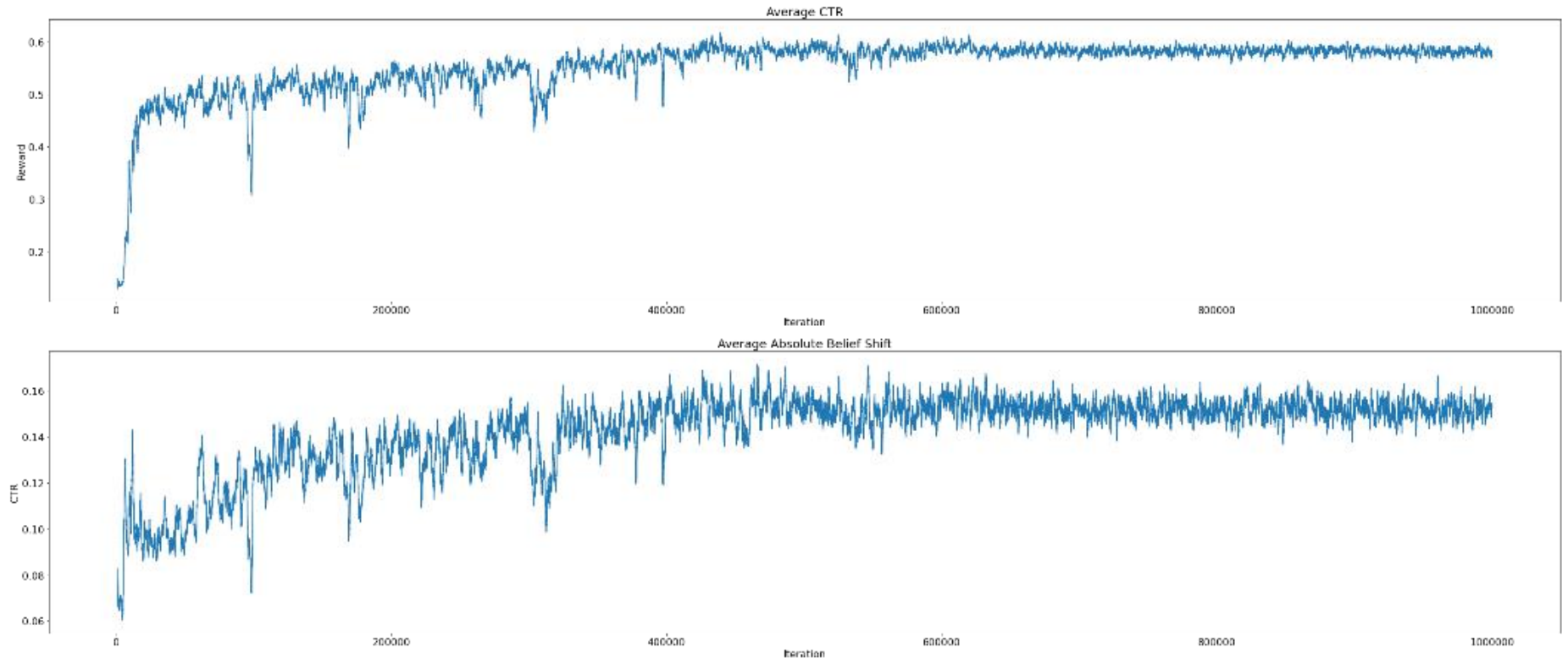

Figure 5.15: Depolarization agent: training metrics

Also similar to the polarization model, the depolarization model produced a dramatic shift in the belief distribution of the test population. The spread of belief is much smaller, and two large peaks are present around -0.25 and 0.25 on the belief spectrum with a slight peak at belief scores of 0.

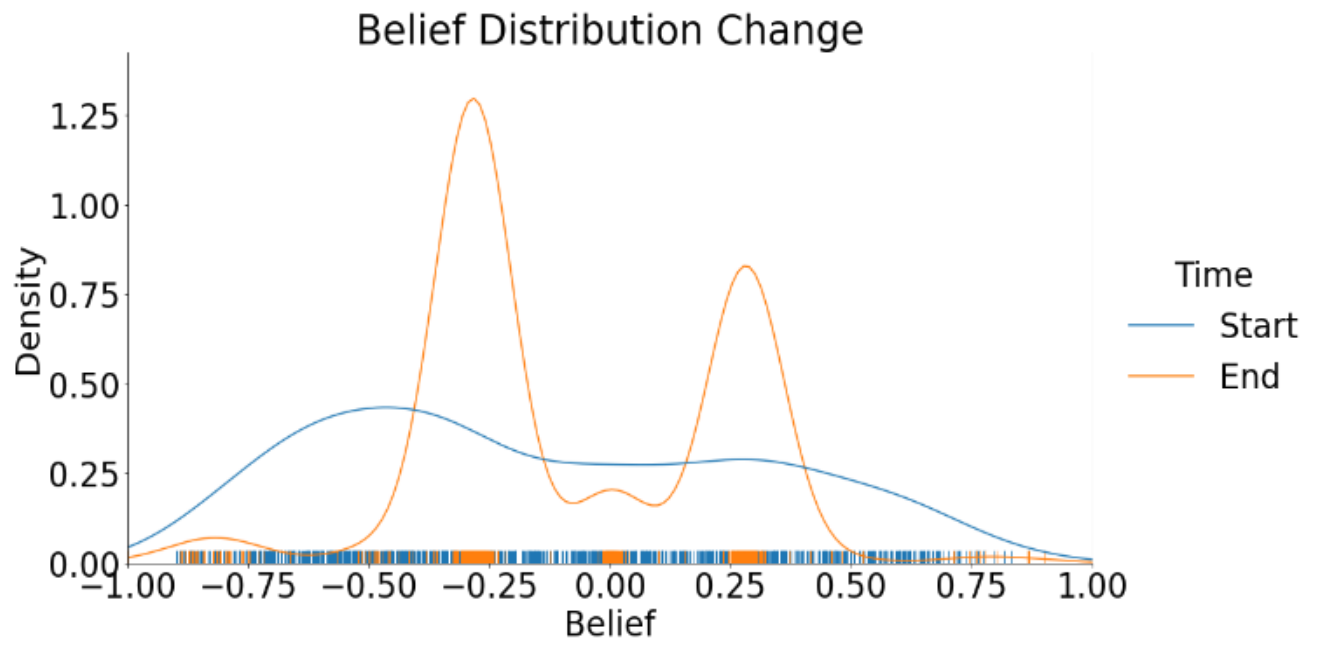

Figure 5.16: Depolarization agent: belief distribution change

Shown in both Figure 5.16 and 5.17, the depolarization model condensed user belief to three regions. We can see that of the three, the center region is the smallest. This is likely due to the depolarization model does not need to shift users anymore towards the center once they are within the region in which lean left or lean right content appeals to them.

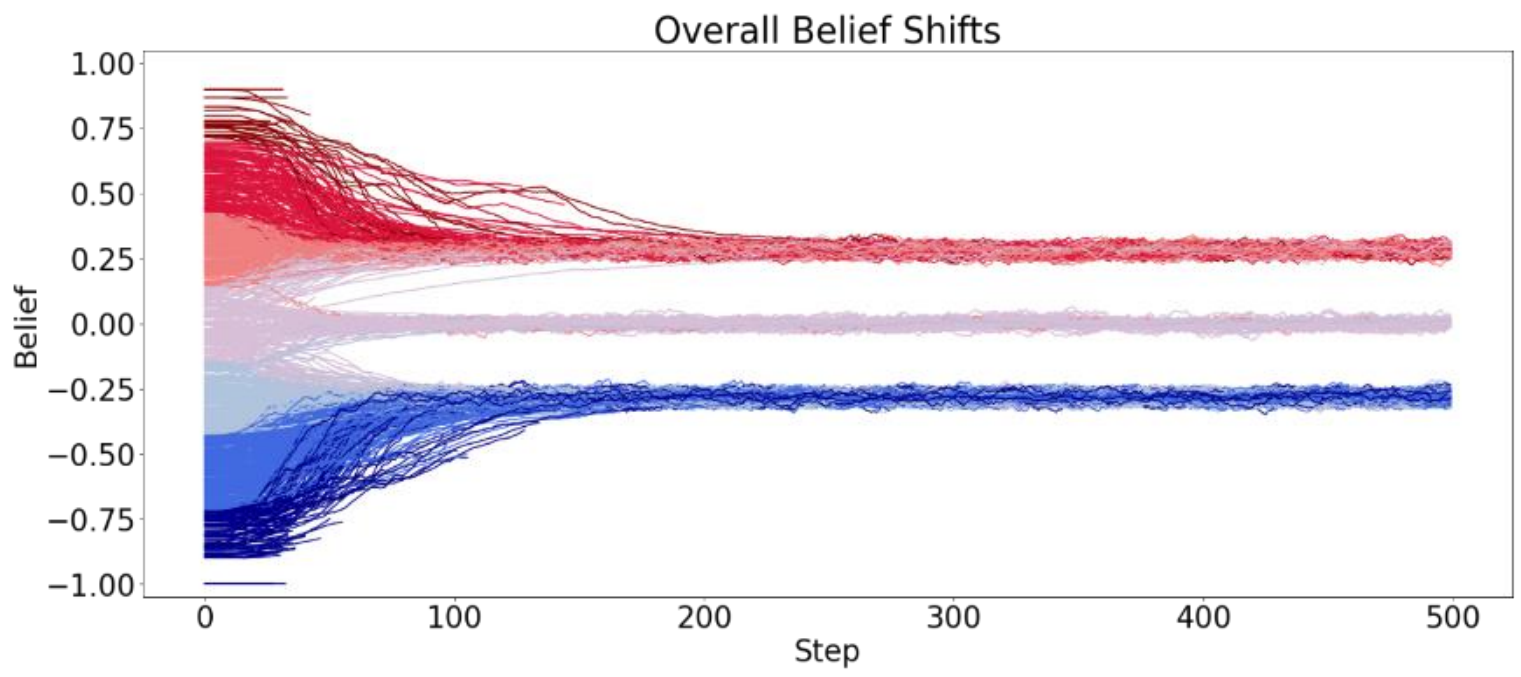

Figure 5.17: Depolarization agent: overall belief shifts

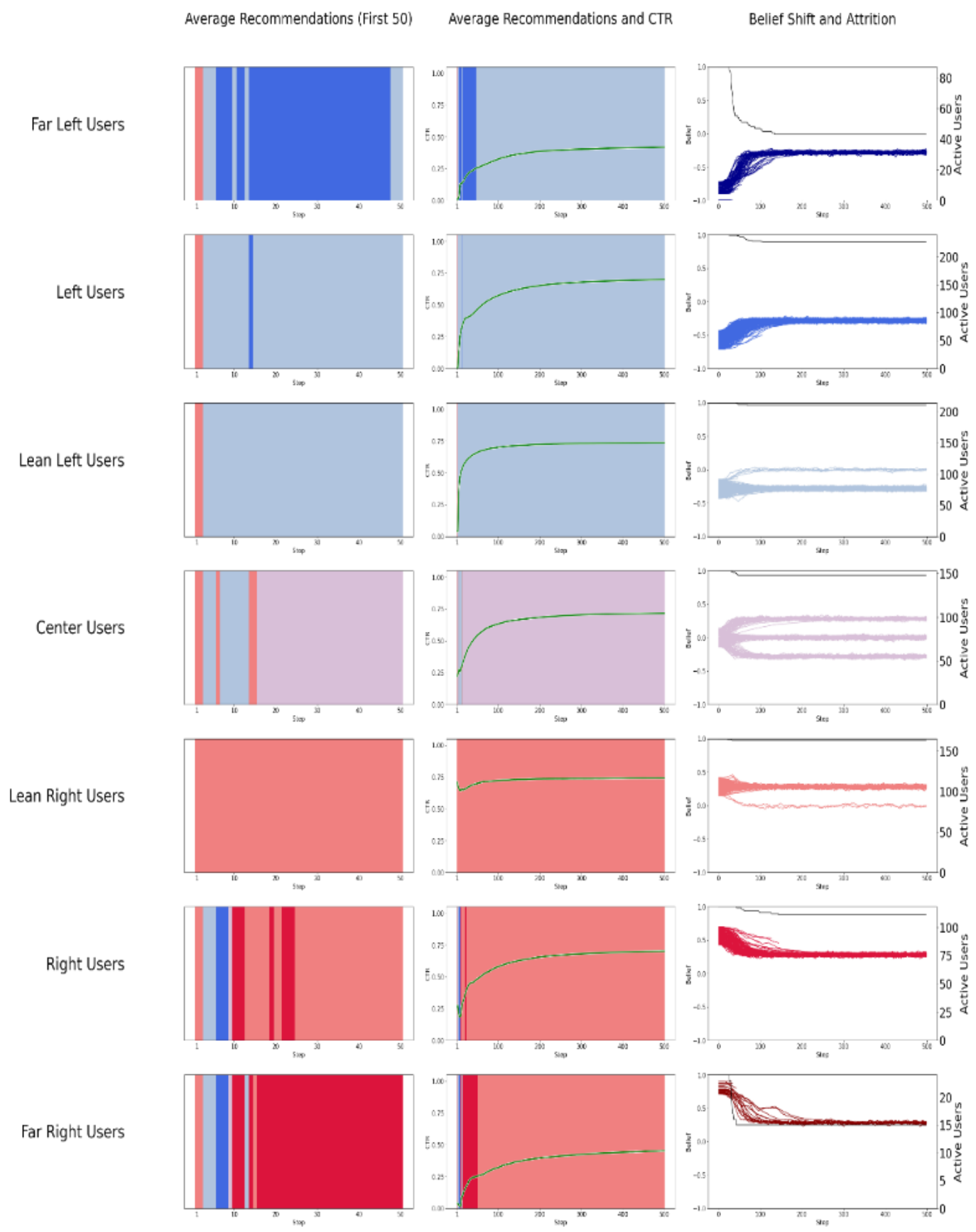

Figure 5.18: Depolarization agent: average recommendations by user ideology type

In Figure 5.18, the policy of the depolarization can be seen to have struggled most with far left and far right users. As with the polarization model and users near the center, the further user belief must shift, the more difficult it is to do so without the user's *satisfaction* dropping and the user leaving the platform.

Unlike the polarization model, the overshooting behavior is not seen for the depolarization model. User belief on average stops shifting after reaching the two outer thresholds around -0.25 and 0.25.

Looking at the average content recommended to far right users, we can see that the depolarization model has learn a strategy of first profiling the user – denoted by recommended a sampling of different content initially – and then recommending right content until the user is likely enough to click on lean right content. This gradual shifting behavior is also observed for far left users.

## 5.6 Comparison

To directly compare the performance of each agent, metrics for CTR and attrition rate can be found in Table 5.1.

| Agent | Metrics | |
|---|---|---|
| | CTR | Attrition Rate |
| Random | 0.17 | 1.00 |
| Baseline | 0.61 | 0.20 |
| No Manipulation | 0.60 | 0.15 |
| Polarization | 0.59 | 0.18 |
| Depolarization | 0.51 | 0.23 |

Table 5.1: Comparison of CTR and attrition rates for each agent

The best performing agent in terms of CTR was the baseline agent while the best performing agent in terms of attrition rate was the no manipulation agent. Likely, the baseline agent spends more time than necessary profiling users. In this way, it's perhaps more certain of its recommendations but ends up causing more users to attrit. The no manipulation model on the other hand more quickly profiles users, leading to a lower attrition rate. However due to not recommending far left or far right content, the no manipulation agent ends up with less clicks for far left and far right users.

Regarding the agents with manipulative reward functions, both performed worse on the two metrics. This makes sense in that these agents must spend time shifting belief before a user is receptive to the target content. This causes an increased risk of attrition and lower CTR overall. These are in no doubt a strong argument against the case of deploying such manipulative agents in the real world. Increasing customer churn and having lower overall engagement may not be worth the influence such agents would have on users. Still, it's possible with more training, different RL algorithms, and/or more complicated and nuanced reward functions, these risks could be better mitigated.

Interestingly it appears that it was more difficult to depolarize users than to polarize them. This is likely due to the depolarization agent being unable to exploit the repulsive effect of opposing content. On the other hand, the polarization agent can use both the attractive and repulsive forces.

Overall, these simple experiments demonstrate that, given the described dynamics of political opinion shift on social news and media platforms, simulated environments can allow reinforcement learning agents to learn policies that manipulate user preferences according to a given intentional aim.

# 6 | Conclusion

## 6.1 Summary

In this paper, the intersection of the fields of reinforcement learning based recommender systems, social news and media platforms, and political opinion dynamics was studied and analyzed through review of research and literature. It was reasoned that advancements in understanding and technology in each of these areas lead to the emergence of a threat of particular concern: purposeful manipulation of the belief and opinions of users on social media platforms. The risks and detrimental effects of recommender system have historically been highlighted, especially the ability of recommender systems to shift user preferences in undesirable ways. Mitigation strategies have been and continue to be developed for such concerns. However, much less study has been done on the ways in which bad actors could intentionally tamper with recommender systems to capitalize on this ability.

To shed light on this concern, experiments were conducted which demonstrate how this could be accomplished and along with its feasibility. This was done by simulating a social news and media environment which incorporated known and studied dynamics of how opinions may shift when exposed to political content. User belief and content bias were both latent features corresponding to the ideological opinions of users or the political partiality of content (i.e., news articles). This environment allowed for the shifting of user belief in either direction on a continuous, 1-deminsional scale when exposed to various content bias which also exists on the same scale. Reinforcement learning agents in this environment however were not privy to these latent features but instead were only given previous recommendations and whether the user clicked on the recommended content as noisy observations.

It was shown that by simply modifying the reward function of agents, policies which shifted user belief in a desired manner. The agent in one experiment was only rewarded when a user clicked on extreme content while another agent was only reward when a user clicked on more neutral content. In both cases, the agents were easily able to learn how to manipulate the latent user features to maximize rewards across subsequent recommendations.

This simulated environment was only pseudo-realistic and relied on many assumptions. It also did not incorporate any real user data which would be used to create such a simulation for a real reinforcement learning based recommender system. User data may never be perfect and complete however and the signal in a real environment would be far noisier. Thus, the results of these experiments are not empirical. Although real, observed properties of political opinion dynamics were incorporated, there are likely magnitudes more factors at play.

However, the argument can be made that as political opinion dynamics modeling improves and companies collect increasingly more user data, it stands to reason that we will approach near-perfect simulations of such environments at some point in the future. Thus, given the demonstrated feasibility, this paper calls for research into methods of detection and mitigation for recommender systems which are intentionally incentivized to manipulate users.

## 6.2 Future Work

The simulated environment designed for this thesis could be improved and expanded in several areas. These include:

- Using real user data to set the user feature parameters (*polarization factor*, *malleability*, etc.)
- Using real interaction and engagement data to better model click probabilities
- Adding additional user features
- Incorporating a mechanism in which user belief may randomly shift due to forces outside recommendations
- Using graph theory to incorporate social network properties (i.e., users can follow other users)
- Make content be "produced" by users according to their belief

Additionally, further experimentation could be done which includes:

- Increasing noise (lower click probabilities, smaller belief shifts) to see when explicit manipulation becomes infeasible due to computational restraints
- Experiment with different *horizon* and *user lifespan* lengths
- Use other RL algorithms such as policy gradient methods
- Use RNN or LSTM layers to better handle the sequential observations

The components of the simulation environment and future improvements can be found on GitHub - https://github.com/matthewsparr/POD-SNMPs.

# References

1. Adam D. I. Kramer, Jamie E. Guillory, and Jeffrey T. Hancock. Experimental evidence of massive scale emotional contagion through social networks. National Academy of Sciences, 111(24):8788 –8790. (2014). https://www.pnas.org/content/111/24/8788.

2. Adomavicius, Gediminas & Bockstedt, Jesse & Curley, Shawn & Zhang, Jingjing. Recommender Systems, Consumer Preferences, and Anchoring Effects. CEUR Workshop Proceedings. 811. (2011). https://citeseerx.ist.psu.edu/viewdoc/download?doi=10.1.1.423.3360&rep=rep1&type=pdf.

3. Aggarwal, Charu C. Recommender Systems. Springer International Publishing, 2016. DOI.org (Crossref), https://doi.org/10.1007/978-3-319-29659-3.

4. Bail, Christopher A., et al. "Exposure to Opposing Views on Social Media Can Increase Political Polarization." Proceedings of the National Academy of Sciences, vol. 115, no. 37, Sept. 2018, pp. 9216–21. DOI.org (Crossref), https://doi.org/10.1073/pnas.1804840115.

5. Becker, Joshua, et al. "The Wisdom of Partisan Crowds." Proceedings of the National Academy of Sciences, vol. 116, no. 22, 2019, pp. 10717–22, https://doi.org/10.1073/pnas.1817195116.

6. Blog, Netflix Technology. "Artwork Personalization at Netflix." Medium, 7 Dec. 2017, https://netflixtechblog.com/artwork-personalization-c589f074ad76.

7. Brooks, Heather Z., and Mason A. Porter. "A Model for the Influence of Media on the Ideology of Content in Online Social Networks." Physical Review Research, vol. 2, no. 2, Apr. 2020, https://doi.org/10.1103/physrevresearch.2.023041.

8. Chen, Xinshi, et al. Generative Adversarial User Model for Reinforcement Learning Based Recommendation System. (2020). https://arxiv.org/pdf/1812.10613.pdf.

9. Clark, Jack, and Dario Amodei. Faulty reward functions in the wild. (2016). https://blog.openai.com/faulty-reward-functions/.

10. David Rohde, Stephen Bonner, Travis Dunlop, Flavian Vasile, and Alexandros Karatzoglou. Recogym: A reinforcement learning environment for the problem of product recommendation in online advertising. (2018). https://arxiv.org/pdf/1808.00720.pdf.

11. Del Vicario, M., Scala, A., Caldarelli, G. et al. Modeling confirmation bias and polarization. Sci Rep 7, 40391 (2017). https://doi.org/10.1038/srep40391.

12. D. Knopoff, "On a mathematical theory of complex systems on networks with application to opinion formation," Mathematical Models and Methods in Applied Sciences 24, 405–426 (2014).

13. Eugene Ie, Chih-wei Hsu, Martin Mladenov, Vihan Jain, Sanmit Narvekar, Jing Wang, Rui Wu, and Craig Boutilier. Recsim: A configurable simulation platform for recommender systems. (2019). https://arxiv.org/pdf/1909.04847.pdf.

14. Evans, Charles, and Atoosa Kasirzadeh. "User Tampering in Reinforcement Learning Recommender Systems." (2021). http://arxiv.org/abs/2109.04083.

15. Guess, Andrew M., et al. "The Consequences of Online Partisan Media." Proceedings of the National Academy of Sciences, vol. 118, no. 14, 2021, https://doi.org/10.1073/pnas.2013464118.

16. Gulcehre, Caglar, et al. Addressing Extrapolation Error in Deep Offline Reinforcement Learning. (2021). https://openreview.net/forum?id=OCRKCul3eKN.

17. Gupta, Abhishek, et al. "Relay Policy Learning: Solving Long-Horizon Tasks via Imitation and Reinforcement Learning." ArXiv:1910.11956 [Cs, Stat], Oct. 2019. arXiv.org, http://arxiv.org/abs/1910.11956.

18. Hare, Christopher, and Keith T Poole, "The polarization of contemporary American politics," Polity 46, 411–429 (2014). https://www.journals.uchicago.edu/doi/pdf/10.1057/pol.2014.10.

19. Hausknecht, Matthew, and Peter Stone. Deep Recurrent Q-Learning for Partially Observable MDPs. (2017). https://arxiv.org/pdf/1507.06527.

20. Horgan, Dan, et al. Distributed Prioritized Experience Replay. (2018). https://arxiv.org/pdf/1803.00933.

21. Ie, Eugene, et al. "RecSim: A Configurable Simulation Platform for Recommender Systems." ArXiv:1909.04847 [Cs, Stat]. (2019). http://arxiv.org/abs/1909.04847.

22. Jan Kleinnijenhuis, Anita M J van Hoof, Wouter van Atteveldt, The Combined Effects of Mass Media and Social Media on Political Perceptions and Preferences, Journal of Communication, Volume 69, Issue 6, December 2019, Pages 650–673, https://doi.org/10.1093/joc/jqz038

23. Kottonau, Johannes & Pahl-Wostl, Claudia. Simulating political attitudes and voting behavior. Journal of Artificial Societies and Social Simulation. 7. (2004). https://www.researchgate.net/profile/Johannes-Kottonau/publication/5140405_Simulating_political_attitudes_and_voting_behavior/links/574ff62608ae10b2ec0568c3/Simulating-political-attitudes-and-voting-behavior.pdf.

24. Krueger, David, Tegan Maharaj, and Jan Leike. Hidden Incentives for Auto-Induced Distributional Shift. arXiv:2009.09153 [cs, stat]. (2020). URL http://arxiv.org/abs/2009.

25. Kurniawati, Hanna. "Partially Observable Markov Decision Processes (POMDPs) and Robotics." ArXiv:2107.07599 [Cs], July 2021. arXiv.org, http://arxiv.org/abs/2107.07599.

26. Levy, R. Social Media, News Consumption, and Polarization: Evidence From a Field Experiment. American Economic Review. (2021). https://www.aeaweb.org/articles/pdf/doi/10.1257/aer.20191777

27. Liang, Eric, et al. RLlib: Abstractions for Distributed Reinforcement Learning. (2018). https://arxiv.org/pdf/1712.09381.

28. Liu, Evan Zheran, et al. Learning Abstract Models for Long-Horizon Exploration. (2019). https://openreview.net/forum?id=ryxLG2RcYX.

29. M. Mehdi Afsar, Trafford Crump, and Behrouz Far. Reinforcement learning based recommender systems: A survey. arXiv:2101.06286 [cs]. (2021). http://arxiv.org/abs/2101.06286.

30. Meshram, Rahul, et al. "Optimal Recommendation to Users That React: Online Learning for a Class of POMDPs." ArXiv:1603.09233 [Cs], Mar. 2016. arXiv.org, http://arxiv.org/abs/1603.09233.

31. Michela Del Vicario, Antonio Scala, Guido Caldarelli, H Stanley, and Walter Quattrociocchi, "Modeling confirmation bias and polarization," Scientific Reports 7, 40391 (2016). https://www.nature.com/articles/srep40391.pdf.

32. Mnih, Volodymyr, et al. Playing Atari with Deep Reinforcement Learning. (2013). https://arxiv.org/pdf/1312.5602.pdf.

33. Mladenov, Martin, et al. Advantage Amplification in Slowly Evolving Latent-State Environments. (2019). https://arxiv.org/pdf/1905.13559.pdf.

34. Nyhan, B.; and Reifler, J. When Corrections Fail: The Persistence of Political Misperceptions. Political Behavior. (2010).

35. Pew Research Center, June 2016, "Partisanship and Political Animosity in 2016". https://www.pewresearch.org/politics/wp-content/uploads/sites/4/2016/06/06-22-16-Partisanship-and-animosity-release.pdf.

36. Pew Research Center, October 2017, "The Partisan Divide on Political Values Grows Even Wider. https://www.pewresearch.org/politics/wp-content/uploads/sites/4/2017/10/10-05-2017-Political-landscape-release-updt..pdf.

37. Pew Research Center, January 2021, "News Use Across Social Media Platforms in 2020". https://www.pewresearch.org/journalism/wp-content/uploads/sites/8/2021/01/PJ_2021.01.12_News-and-Social-Media_FINAL.pdf.

38. Raza, Shaina, and Chen Ding. "News Recommender System: A Review of Recent Progress, Challenges, and Opportunities." (2021). http://arxiv.org/abs/2009.04964.

39. Sabin-Miller, D., & Abrams, D. M. When pull turns to shove: A continuous-time model for opinion dynamics. Physical Review Research, 2(4). (2020). https://arxiv.org/pdf/2005.05478.

40. S. C. Matz, M. Kosinski, G. Nave, and D. J. Stillwell. Psychological targeting as an effective approach to digital mass persuasion. Proceedings of the National Academy of Sciences, 114(48): 12714–12719. (2017) http://www.pnas.org/lookup/doi/10.1073/pnas.1710966114.

41. S. Sharma, A. S. Lakshminarayanan, and B. Ravindran. Learning to repeat: Fine-grained action repetition for deep reinforcement learning. ICLR-17, Toulon, France. (2017). https://openreview.net/pdf?id=B1GOWV5eg.

42. Silvia Milano, Mariarosaria Taddeo, and Luciano Floridi. Recommender systems and their ethical challenges. AI & SOCIETY 35, 4 (2020). https://link.springer.com/content/pdf/10.1007/s00146-020-00950-y.pdf.

43. Stray, Jonathan. "Designing Recommender Systems to Depolarize." ArXiv:2107.04953 [Cs], July 2021. arXiv.org, http://arxiv.org/abs/2107.04953.

44. Stuart Armstrong, Jan Leike, Laurent Orseau, and Shane Legg. 2020. Pitfalls of Learning a Reward Function Online. In Proceedings of the Twenty-Ninth International Joint Conference on Artificial Intelligence, IJCAI-20, Christian Bessiere (Ed.). International Joint Conferences on Artificial Intelligence Organization, 1592–1600. (2020). https://arxiv.org/pdf/2004.13654.pdf.

45. Sutton, Richard S., and Andrew G. Barto. Reinforcement Learning: An Introduction. Second edition, The MIT Press, 2018.

46. Tom Everitt, Marcus Hutter, Ramana Kumar, and Victoria Krakovna. 2021. Reward tampering problems and solutions in reinforcement learning: A causal influence diagram perspective. Synthese. (2021) https://arxiv.org/pdf/1908.04734.pdf.

47. Ulrike Gretzel and Daniel Fesenmaier. Persuasion in Recommender Systems. International Journal of Electronic Commerce, 11(2):81–100. (2006). https://doi.org/10.2753/JEC1086-4415110204.

48. Wang, Kai, et al. RL4RS: A Real-World Benchmark for Reinforcement Learning Based Recommender System. (2021). https://arxiv.org/pdf/2110.11073.pdf.

49. Wang, Lu, et al. Supervised Reinforcement Learning with Recurrent Neural Network for Dynamic Treatment Recommendation. (2018). https://arxiv.org/pdf/1807.01473.pdf.

50. Zheng, Guanjie, et al. “DRN: A Deep Reinforcement Learning Framework for News Recommendation.” Proceedings of the 2018 World Wide Web Conference, International World Wide Web Conferences Steering Committee, 2018, pp. 167–76, https://doi.org/10.1145/3178876.3185994.

51. Zhongqi, Lu and Qiang Yang. Partially observable markov decision process for recommender systems. (2016). https://arxiv.org/pdf/1608.07793.

52. Anonymous. “Estimating and Penalizing Induced Preference Shifts in Recommender Systems.” Submitted to The Tenth International Conference on Learning Representations. (2022). https://openreview.net/forum?id=kiNEOCSEzt.